%% file: arxiv_submission.tex
\documentclass[10pt,twocolumn,letterpaper]{article}

\usepackage{iccv}
\usepackage{times}
\usepackage{epsfig}
\usepackage{graphicx}
\usepackage{amsmath}
\usepackage{amssymb}
\usepackage{caption}
\usepackage{multirow}
\usepackage{pifont}
\usepackage[dvipsnames]{xcolor}
\usepackage{rotating}
\usepackage{makecell}
\usepackage{booktabs} 
\usepackage[accsupp]{axessibility}  

\newcommand{\cmark}{\textcolor{green}{\ding{51}}}%
\newcommand{\xmark}{\textcolor{red}{\ding{55}}}%

\newcommand{\gwd}[0] {Gromov-Wasserstein Discrepancy}

\definecolor{somegray}{rgb}{0.5, 0.5, 0.5}
\newcommand{\darkgrayed}[1]{\textcolor{somegray}{#1}}
\makeatletter
\newcommand*\titleheader[1]{\gdef\@titleheader{#1}}
\AtBeginDocument{%
  \let\st@red@title\@title
  \def\@title{%
    \vskip-6em
    \bgroup\normalfont\large\centering\@titleheader\par\egroup
    \vskip1.5em\st@red@title}
}
\makeatother

\titleheader{\darkgrayed{This paper has been accepted for publication at the\\
IEEE/CVF International Conference on Computer Vision (ICCV), Paris, 2023.
\copyright IEEE}}

\title{From Chaos Comes Order: \\Ordering Event Representations for Object Recognition and Detection}

\usepackage[pagebackref=true,breaklinks=true,letterpaper=true,colorlinks,bookmarks=false]{hyperref}

\iccvfinalcopy 

\def\iccvPaperID{10936} 

\ificcvfinal\fi

\begin{document}

\makeatletter
\let\@oldmaketitle\@maketitle
\renewcommand{\@maketitle}{\@oldmaketitle
\centering
\setlength{\tabcolsep}{2pt} 
\begin{tabular}{cc}
\includegraphics[height=4cm]{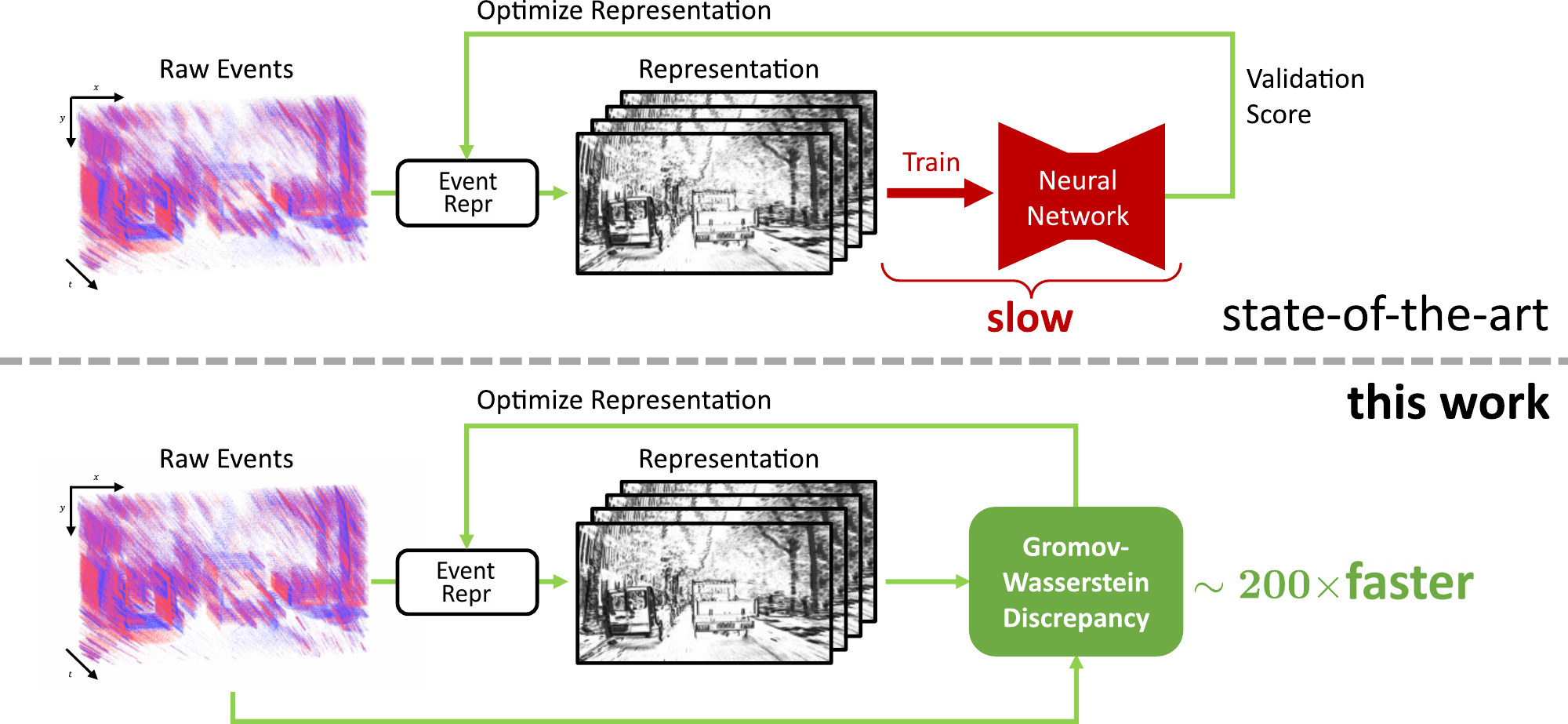}&
\includegraphics[height=4cm]{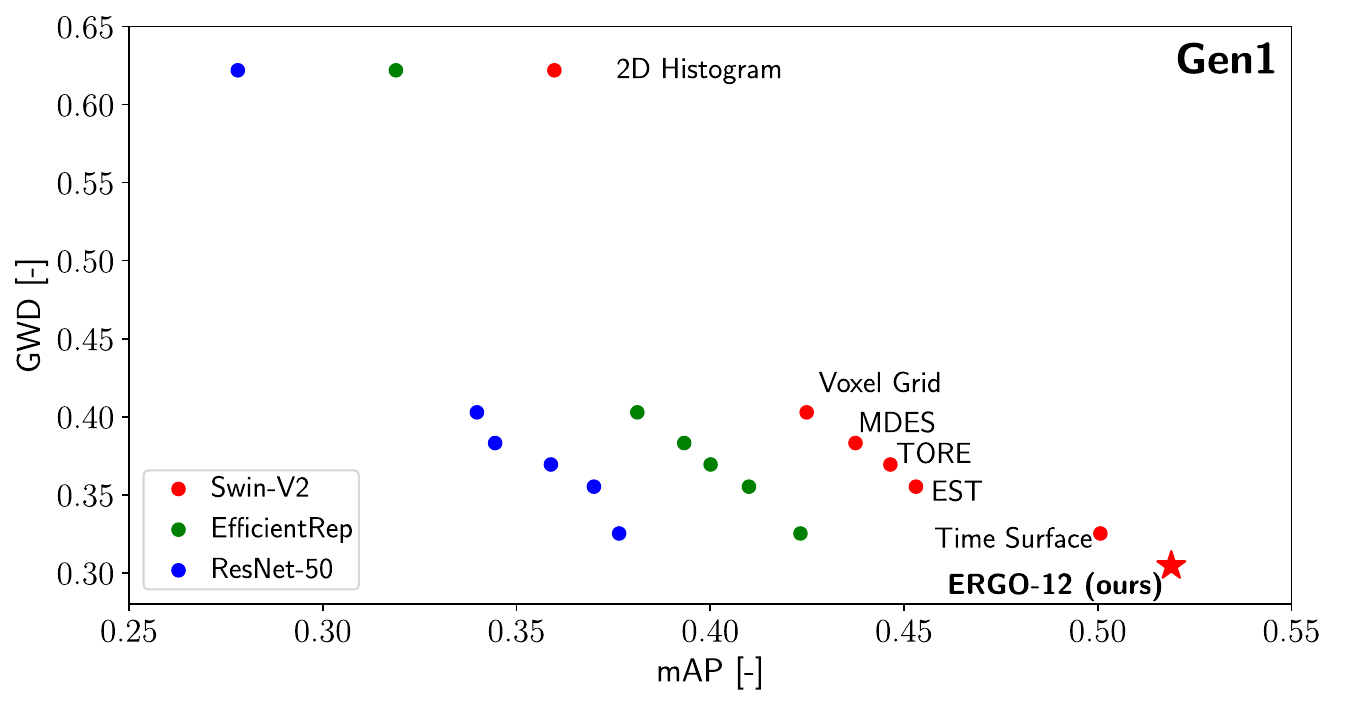}\\
(a) Method overview & (b) GW Discrepancy vs. Performance
\end{tabular}

\vspace{-1ex}
\captionof{figure}{Selecting dense event representations for deep neural networks is exceedingly slow since it involves training a neural network for each representation and selecting the best one based on the validation score. In this work, we eliminate this bottleneck by selecting the representation based on the \gwd~ (GWD) on the validation set (a). This metric is $200$ times faster to compute and preserves the task performance ranking of event representations across multiple representations, network backbones, datasets and tasks (b, showing object detection performance on the Gen1~\cite{Tournemire20arxiv} dataset). We use it to, for the first time, perform a hyperparameter search on a large family of event representations, revealing new and powerful event representations that exceed the state-of-the-art.\\}
\label{fig:eyecatcher}
}
\makeatother
\setlength{\tabcolsep}{10pt} 

\author{
    Nikola Zubi\'{c}$^*$
    \qquad
    Daniel Gehrig\thanks{Equal contribution}
    \qquad
    Mathias Gehrig
    \qquad
    Davide Scaramuzza\\
    \vspace*{-10pt}\\
    Robotics and Perception Group, University of Zurich, Switzerland
}
    
\maketitle
\ificcvfinal\thispagestyle{empty}\fi

\begin{abstract}
Today, state-of-the-art deep neural networks that process events first convert them into dense, grid-like input representations before using an off-the-shelf network. However, selecting the appropriate representation for the task traditionally requires training a neural network for each representation and selecting the best one based on the validation score, which is very time-consuming. This work eliminates this bottleneck by selecting representations based on the \gwd~(GWD) between raw events and their representation. It is about $200$ times faster to compute than training a neural network and preserves the task performance ranking of event representations across multiple representations, network backbones, datasets, and tasks. Thus finding representations with high task scores is equivalent to finding representations with a low GWD. We use this insight to, for the first time, perform a hyperparameter search on a large family of event representations, revealing new and powerful representations that exceed the state-of-the-art. Our optimized representations outperform existing representations by 1.7 mAP on the 1 Mpx dataset and 0.3 mAP on the Gen1 dataset, two established object detection benchmarks, and reach a 3.8\% higher classification score on the mini N-ImageNet benchmark. Moreover, we outperform state-of-the-art by 2.1 mAP on Gen1 and state-of-the-art feed-forward methods by 6.0 mAP on the 1 Mpx datasets. This work opens a new unexplored field of explicit representation optimization for event-based learning.  
\end{abstract}

\section*{Multimedia Material}
For open-source code please visit \url{https://github.com/uzh-rpg/event_representation_study}.

\section{Introduction}
Event cameras are biologically inspired vision sensors that function in a fundamentally distinct way \cite{Gallego20pami}. Unlike traditional cameras that capture images at a fixed rate, these cameras measure \textit{brightness changes} independently for each pixel, and these changes are referred to as events. The events encode the time, location, and polarity (sign) of the brightness changes. Event cameras offer several advantages over frame-based cameras, including exceptionally high temporal resolution (in the order of $\mu$s), a high dynamic range, and low power consumption. Their numerous benefits make them attractive for a wide range of applications like robotics, autonomous vehicles, and virtual reality. However, due to their sparse and asynchronous nature, applying classical computer vision algorithms remains challenging.

Many state-of-the-art deep learning models address this challenge by converting sparse and asynchronous events into dense grid-like representations before processing them with off-the-shelf deep neural networks. By using these networks, methods like this enjoy the advantages of mature learning algorithms and network architectures, and optimized hardware, but we need to make a non-trivial choice of event representation.
In fact, the computer vision and robotics fields are witnessing a surge in the number of research papers utilizing event-based vision, resulting in a plethora of new event representations being proposed. Despite this, extensive comparisons of these representations remain rare, making it unclear whether these newer representations should be adopted.

No efficient methodology exists for comparing these representations.
Conventionally, comparing event representations involves training a fixed deep-learning model for each event representation separately and subsequently selecting the optimal one based on a validation score. This process is very time-intensive since it requires network training in the loop which often takes hours or days~(Fig.~\ref{fig:eyecatcher} a top). 

In this work, we propose a fast method to compare event representations which circumvents the need to train a neural network and instead computes the \textit{\gwd~(GWD)} between the raw events and event representation~(Fig.~\ref{fig:eyecatcher} a bottom). 
This metric effectively measures the distortion that is introduced through converting raw events to representations and thus puts an upper bound on the amount of information that can be accessed by downstream neural networks. We show extensive experimental evidence, that this metric preserves the task-performance ranking across a wide range of input representations for several datasets, neural network backbones and tasks (Fig.~\ref{fig:eyecatcher} b).    
Due to its low computational cost, we apply the GWD to, for the first time, explicitly optimize over a large family of event representations, which reveals a new and powerful representation, which we term 12-channel \textbf{E}vent \textbf{R}epresentation through \textbf{G}romov-Wasserstein \textbf{O}ptimization (ERGO-12). For the task of object detection, networks trained with these representations outperform other representations by 1.9 mAP on the 1 Mpx dataset and 0.3 mAP on Gen1, even outperforming state-of-the-art methods by 2.1 mAP on Gen1 and state-of-the-art feed-forward methods by 6.0 mAP on the 1 Mpx dataset. On object recognition, we instead find that our representation outperforms state-of-the-art representations by 3.8\%. We believe that the GWD is a powerful tool that opens up a new research field that searches for optimized event representations. Our contributions are summarized as follows:
\begin{itemize}
  \setlength{\itemsep}{0pt}
  \item We introduce a novel, efficient approach for comparing dense event representations using the \textit{\gwd~(GWD)}. 
  \item We show extensive experimental evidence that it preserves the task performance ranking of neural networks trained with these representations across datasets, neural network backbones and tasks.
  \item We use it to, for the first time, conduct a hyperparameter search on a vast family of event representations, unveiling novel and powerful event representations that outperform the current state-of-the-art representations on the object detection and object classification task.
\end{itemize}
\section{Related Work}
In the field of event-based vision, two primary groups of representations exist: sparse and dense. Methods that use sparse representations \cite{Orchard15pami, Zhao15tnnls, Lee16fns, PerezCarrasco13pami} preserve the sparsity in the events but do not yet scale to more complex tasks due to a lack of specialized hardware and mature neural networks architectures. This frequently results in lower performance on downstream tasks. In contrast, dense representations \cite{Maqueda18cvpr, Sironi18cvpr, Zhu19cvpr, Wang19cvpr} offer improved performance since they can leverage mature machine learning algorithms and neural network architectures. 
\input{figures_and_tables/method_overview.tex}
Sparse representations, pioneered by asynchronous SNNs \cite{Orchard15pami, Lee16fns, PerezCarrasco13pami}, are limited by the lack of specialized hardware and computationally efficient backpropagation algorithms. Point cloud encoders \cite{Sekikawa19cvpr, Qi17nips, fan2021pstnet} have been used due to the spatio-temporal nature of event data, but can be computationally expensive and noisy. Graph neural networks \cite{Li21iccv, Schaefer22cvpr, Bi20trip, bi19iccv, mondal2021moving, Deng22cvpr} are scalable and have achieved high performance on various vision tasks but are still less accurate than dense methods for event-based vision. In this study, we focus on dense event representations and aim to achieve better task performance by utilizing existing efficient learning algorithms that are appropriate for current hardware.

Early dense representations converted events to histograms~\cite{Maqueda18cvpr}, generated time surfaces \cite{Sironi18cvpr} or combined both~\cite{Zhu18rss} while relying on standard neural network backbones to process them. However, these representations only capture a low-dimensional representation of events since they typically only use a few channels. Later approaches have tried to capture more event information by either computing higher-order moments~\cite{Alonso19cvprw} or stacking multiple time windows of events~\cite{Zhu19cvpr}. These methods still stack events based on fixed time windows which is problematic when the event rate becomes too large or too small and lead to the introduction of stacking based on the number of events~\cite{Wang19cvpr}. In parallel, a bio-inspired approach led to the introduction of Time Ordered Recent Event Volumes (TORE) \cite{Baldwin2021TimeOrderedRE}, which aggregate events into queues. However, they are slow to compute and perform similarly to existing Voxel Grids~\cite{Zhu19cvpr}. Most recently, a powerful representation was proposed by Nam et al. \cite{Nam_2022_CVPR}, which divides events into multiple overlapping windows that halve the number of events at each stage, which are more robust during varying scene dynamics. 

Few papers study the effect of event representations on task performance. While~\cite{Perot20nips} and ~\cite{Kim_2021_ICCV} show small-scale ablation studies to select event representations, only Gehrig et al. \cite{Gehrig19iccv} performed a large-scale investigation of event representations by training models on various inputs for multiple tasks. Their study demonstrated the advantages of splitting polarities and incorporating timestamps into representations, and it introduced a learnable representation. However, training for a single task was still computationally expensive, which limited the number of representations that could be compared. For this reason, their study did not cover a large number of representations, and in particular, did not consider different window sizes as is done in later work~\cite{Wang19cvpr, Nam_2022_CVPR}, or more advanced aggregations and measurements like in~\cite{Alonso19cvprw}. Our method instead introduces an efficient metric to compare event representations that solves these limitations, allowing us to perform a search over a large family of event representations and go beyond the representations in \cite{Wang19cvpr,Alonso19cvprw,Nam_2022_CVPR,Gehrig18eccv}, including even non-differentiable hyperparameters.
\section{Method}
\label{sec:method}
In this section, we will first introduce the preliminaries on computing event representations (Sec.~\ref{sec:method:preliminaries}) and then propose the metric we use to measure the discrepancy between events and their representation based on the GWD~(Sec.~\ref{sec:method:GW_discrepancy}) before concluding with Sec.~\ref{sec:method:bayesian_optimization} where we use Bayesian optimization to find an optimal event representation.

\subsection{Preliminaries} 
\label{sec:method:preliminaries}
Event cameras measure brightness changes as an asynchronous stream of \textit{events}. Each event is triggered when the intensity $L(\textbf{u})$ at the pixel $\textbf{u}=(x,y)$ changes by the contrast threshold $C$ at time $t$, and thus satisfies
\begin{equation}
    p[L(\textbf{u},t)-L(\textbf{u},t-\Delta t)] = C
\end{equation}
where $p\in\{-1,1\}$ is the polarity of the event, and $t-\Delta t$ is the time of the last event.
Within a time window $\Delta T$, an event camera thus generates an ordered set of events $\mathcal{E}=\{e_k\}_{k=0}^{N_e-1}$, with each event $e_k = (\mathbf{u}_k, t_k, p_k)\in \mathbb{R}^4$. To bridge the gap between asynchronous events and dense neural networks, they are usually converted to a dense event representation 
\begin{equation}
    \mathcal{R} = F(\mathcal{E}),
\end{equation}
which has features $f_\mathbf{x}\doteq \mathcal{R(\textbf{x})} \in \mathbb{R}^{N_f}$ indexed by the integer-valued pixel location $\textbf{x}$. The above representation thus generates a set of features $\mathcal{F} = \{f_\mathbf{x}\}_{\mathbf{x}\in\Omega}$, where $\Omega$ denotes the image domain and has size $\vert \Omega\vert= N_f$ with $N_f$ being the number of pixels in the image. In what follows, we will derive a measure to quantify the distortion between events $\mathcal{E}$ and features $\mathcal{F}$ based on the GWD.

\subsection{\gwd}
\label{sec:method:GW_discrepancy}

Converting raw events to a representation invariably distorts the events by removing important distinguishing features from the stream. We would like to measure this distortion since we expect it correlates strongly with a neural network's ability to extract features from these events. In what follows, we will derive a measure of this distortion rate based on the GWD. 

We show an overview of the GWD in Fig.~\ref{fig:method_overview}. We start by measuring the similarity between a set of events and their representation by building a soft correspondence between events $e_i$ and features $f_{\textbf{x}_j}$, which we denote as $T_{ij}$\footnote{In the optimal transport literature, this correspondence is also called a \textit{transport plan}.}. This transport plan effectively moves each event to a corresponding feature, thereby distorting the original set and destroying information. 
Importantly, such a plan can be interpreted as follows: We transport the events with a total weight of $1$, i.e. per-event weight $\frac{1}{N_e}$ to the output features, which also need to receive a total weight of $1$ or per-feature weight of $\frac{1}{N_f}$. By this construction, $T_{ij}$ needs to satisfy $\sum_{i}T_{ij} = 1/N_f$ and $\sum_{j}{T_{ij}} = 1/N_e$. This means that a total weight of $1/N_e$ moves from each event, and each feature receives a total weight of $1/N_f$.  

In the next step, we measure the distortion introduced by this transportation plan by considering pairwise similarities of input events and features. Let $e_i,e_k \in\mathcal{E}$ be a pair of events and $f_{\mathbf{x}_j},f_{\mathbf{x}_l} \in\mathcal{F}$ pair of features, with similarity scores $C^e_{ik} \doteq C^e(e_i, e_k)$ and $C^f_{jl}\doteq C^f(f_{\mathbf{x}_j},f_{\mathbf{x}_l})$ between events and features respectively. Next, consider how the transport plan $T$ acts on these pairs: Generally, a weight $T_{ij}$ is moved from event $e_i$ to feature $f_{\mathbf{x}_j}$. Similarly, the weight $T_{kl}$ is moved from event $e_k$ to feature $f_{\mathbf{x}_l}$. Ideally, such a transport plan should preserve the similarity between pairs of source events and target features, and thus the difference in similarity scores between pairs $(i,k)$ and $(j,l)$ can be used as a measure of distortion. For each event pair and feature pair, we define the distortion as 
\begin{equation}
\label{eq:distortion_single_pair}
L_{ijkl} = T_{ij}T_{kl}\mathcal{L}(C^e_{ik}, C^f_{jl}),
\end{equation}
where $\mathcal{L}$ denotes some disparity measurement between $C^e_{ik}$ and $C^f_{jl}$. Summing over all possible pairs of events and features we thus arrive at the transportation cost: 
\begin{equation}
L(T;\mathcal{E}, \mathcal{F}) = \sum_{i,j,k,l}L_{ijkl} = \sum_{i,j,k,l}T_{ij}T_{kl}\mathcal{L}(C^e_{ik}, C^f_{jl})
\end{equation}
Minimizing over transport plans, we arrive at the GWD:
\begin{equation}
\label{eq:GW_discrepancy}
L(\mathcal{E}, \mathcal{F}) = \min_{T} \sum_{i,j,k,l}T_{ij}T_{kl}\mathcal{L}(C^e_{ik}, C^f_{jl}),
\end{equation}
which can be optimized efficiently using \cite{peyre:hal-01322992}.
Since the above metric is defined for a single time window of events, we may average it over multiple samples to find:
\begin{equation}
\label{eq:GW_discrepancy_batched}
\text{GWD}_{N} = \frac{1}{N} \sum_i L(\mathcal{E}_i, \mathcal{F}_i).
\end{equation}
$\text{GWD}_{N}$ can be interpreted as an average distortion rate from raw events to event representations. 
In Sec.~\ref{sec:exp:validation}, we show that GWD$_N$ correlates with a \textit{NN}'s performance with that representation across network backbones, datasets, and event representations. It's also efficient to compute, taking 9 seconds for 50,000 events. In what follows, GWD$_N$ will denote the average over $N$ samples, while GWD  denotes the average over the whole validation set. \\
\textbf{Similarity scores and distortion function} As similarity scores, we choose Gaussian radial basis functions~\cite{SaidSalem2017RGDo} for both events and image features. In detail, 
\begin{align}
C^e_{ik} = e^{\frac{-\Vert e_i - e_k\Vert^2}{2h^2\sigma_e^2}}, \quad C^f_{jl} = e^{\frac{-\Vert f_{\textbf{x}_j} - f_{\textbf{x}_l}\Vert^2}{2h^2\sigma_f^2}}
\end{align}
\begin{align}
    \sigma^2_e &= \underset{i < j}{\text{mean}}\Vert e_i - e_j \Vert^2,\quad
    \sigma^2_f &= \underset{i < j}{\text{mean}}\Vert f_{\textbf{x}_i} - f_{\textbf{x}_j} \Vert^2.
\end{align}
with a bandwidth parameter $h=0.7$.
The selection of data-dependent variances normalizes the distances between pairs of events and features such that the similarity score is robust to the dimensionality of the data and the number of samples in the source and target domain. These details are discussed in \cite{chevallier_dists, SaidSalem2017RGDo, pennec:inria-00614994}. 
While more complex similarity scores could be used, this simple function already achieved good results. As the distortion function, we chose the KL-Divergence i.e. 
\begin{equation}
\label{eq:kl_distortion_function}
    \mathcal{L}(C_{ik}^e,C_{jl}^f) =  C_{ik}^e \log \left(C_{ik}^e/C_{jl}^f\right).
\end{equation}
 As a result, the optimization already rejects terms for which $C_{ik}^e\approx 0$, i.e. pairs of events that are far apart. We found that this property was also beneficial in improving the convergence of the optimization problem in Eq.~\eqref{eq:GW_discrepancy}
\input{figures_and_tables/hyperparameter_overview.tex}\\
\textbf{Improving Convergence of Eq.~\eqref{eq:GW_discrepancy}.}
We found that three features improved convergence and speed up the optimization: (\textit{i}) Normalization of the event coordinates and timestamps by the sensor size and time window respectively, (\textit{ii}) Concatenation of the normalized pixel position to the image features, and (\textit{iii}) Sparsification of image features. Both (\textit{i}) and (\textit{ii}) make the optimization more numerically stable. In fact, without concatenating position information to image features, randomly pixel-wise shuffled event representations would retain the same GWD, although intuitively, neural networks would have a harder time learning from such representations since they typically process nearby features together. Thus reintroducing the position removes this ambiguity and improves the convergence. Finally, (\textit{iii}) removes image features with $\Vert f_{\textbf{x}}\Vert = 0$ since these correspond to pixels where no events were triggered. This step significantly sped up computation by reducing the size of the pair-wise similarity score matrix $C^f$, with a small impact on convergence.
\subsection{Optimizing over Event Representations}
\label{sec:method:bayesian_optimization} 
With a fast method to measure the effectiveness of an event representation, we can now search for the optimal representation by minimizing Eq.~\eqref{eq:GW_discrepancy_batched} over a space of possible representations with a set of hyperparameters $p$. 
\begin{equation}
    p^* =  \arg \min_p \frac{1}{N} \sum_i L(\mathcal{E}_i,\mathcal{F}_i(p)).
\end{equation}
To simplify this optimization, we first describe a very general parametrization, which defines a large family of event representations, extending the family described in~\cite{Gehrig19iccv} in a few ways. These hyperparameters are a small set of categorical variables which can subsequently be optimized using Bayesian optimization.\\
\textbf{Parametrization of Event Representations}
We illustrate the hyperparameters in Fig.~\ref{fig:hyperparameter_overview}. In general, we assume an event representation comprises a stack of features, indexed by $c$ (right side), each of which is derived from (\textit{i}) a specific time window $w_c$ of events, (\textit{ii}) a specific measurement $m_c$ of events such as polarity or timestamps, and (\textit{iii}) a specific aggregation $a_c$, such as summation or averaging. We write such a representation as $N_c$ concatenated feature maps:
\begin{align}
    \mathcal{R}&=\left[\mathcal{R}_0\vert\mathcal{R}_1\vert ...\vert\mathcal{R}_{N_c-1}\right]\\
    \text{with }\mathcal{R}_c &= a_c(m_c(w_c(\mathcal{E})))
\end{align}
Here $w_c$ is a \textit{windowing function} which selects events within an interval, $m_c$ is the \textit{measurement function} which selects an event feature, and $a_c$ is the \textit{aggregation function}, which aggregates measurements into a single feature map. 

Note, in our formulation each channel can have an independent set of parameters, different to~\cite{Gehrig19iccv}, which assumes a shared aggregation and measurement function for all channels. This makes our parametrization substantially more expressive than the one in~\cite{Gehrig19iccv}. The number of (non-learnable) representations is $\left(\vert\mathcal{A}\vert \vert \mathcal{M}\vert \vert \mathcal{W}\vert\right)^{N_c}\approx 3.21\times 10^{27}$, since each channel can be configured independently.
Moreover, while $a_c$ and $m_c$ were already discussed in~\cite{Gehrig19iccv}, the windowing function is a more general concept, illustrated in Fig.~\ref{fig:hyperparameter_overview} (left).
While these windows can be non-overlapping ($w_3$, $w_4$, $w_2$), as for Voxel Grids \cite{Zhu19cvpr}, they can also be overlapping, ($w_0$, $w_1$, $w_2$) as in Mixed-Density Event Stacks~\cite{Nam_2022_CVPR}, or describe windows of a constant event count or constant time~\cite{Wang19cvpr} In this work we allow each feature channel to select from a basis of windows $\mathcal{W} = \{[t_{0,k}, t_{1,k}]\}_{k=0}^{N_k-1}$, which can combine all types of windows unifying these concepts. In summary, a representation is parametrized by: 
\begin{align}
    p=\{(w_c, a_c, m_c)\}_{c=0}^{N_c-1}\\
    \nonumber\text{with}\quad m_c\in \mathcal{M}, a_c\in\mathcal{A}, w_c\in\mathcal{W}.
\end{align}
For the sets of aggregation functions we select $\mathcal{A}~=~\{\text{max}, \text{sum}, \text{mean}, \text{variance}\}$ and for the measurement functions $\mathcal{M}~=~\{t_+, t_-, t, p, c_+, c_-, c\}$ which are most commonly used. Here $c$, $p$, and $t$ denote event count (discarding polarity), polarity, and timestamp. The subscripts $+/-$ select only positive or negative events.
For the basis of time windows, we select three equally spaced, non-overlapping windows from~\cite{Zhu19cvpr} and four overlapping windows from \cite{Nam_2022_CVPR}, including the global window. These are illustrated in Fig.~\ref{fig:exp:bayesian_opt} (right).\\
\textbf{Optimization Procedure} The aforementioned parametrization generates redundant combinations that can be obtained by permuting channels or selecting the same feature for different channels. To address this issue and expedite convergence, we propose a stage-wise optimization procedure. Initially, we start with a volume consisting of zeros with $N_c$ channels and optimize over $a_0, w_0$, and $m_0$ to fill in the first channel. Next, we optimize the feature for the second channel while keeping the first fixed. With this iterative process, we incrementally fill up the representation, avoiding the selection of redundant representations and resulting in faster optimization. At each stage, we use Gryffin \cite{Hase21apr}, a specialized Bayesian optimizer for categorical variables.  



\section{Experiments}
In Sec.~\ref{sec:exp:toy_examples}, we first connect the GWD and event distortion, demonstrating its behavior on several toy examples. Next, in Sec.~\ref{sec:exp:validation}, we showcase the correlation between the metric and task performance across multiple tasks, datasets, backbones, and representations. Lastly, in Sec.~\ref{subsection:optimization_of_event_representations}, we show the outcome of Bayesian Optimization on the GWD and offer insights on the acquired event representation before comparing it against the state-of-the-art.
\subsection{Toy Example}
\label{sec:exp:toy_examples}
As explained in Sec.~\ref{sec:method:GW_discrepancy}, the GWD measures the distortion rate from raw events to event representation. Two experiments were designed to test this claim: (1) analyzing the behavior of the metric when varying the number of bins in Voxel Grid \cite{Zhu19cvpr} and Mixed-Density Event Stack (MDES) \cite{Nam_2022_CVPR} representations, and (2) blurring the event representation with increasing standard deviations before measuring the GWD. We perform experiments on the validation set of Gen1\cite{Tournemire20arxiv} and report the results in Fig.~\ref{fig:exp:toy_examples}.

Fig.~\ref{fig:exp:toy_examples} (top) confirms that the GWD decreases as the number of channels increases in both Voxel Grids and Mixed-Density Event Stacks, which aligns with our intuition that using more channels preserves more information from raw events, resulting in a lower distortion rate. Similarly, the bottom part illustrates that with a growing blur radius, more edges in the event representation are removed, leading to an increase in the GWD. This verifies that the GWD measures the distortion from raw events to event representations.
\input{figures_and_tables/validation_toy_example}

\input{tex/validation_task}
\subsection{Object Detection}
\label{sec:exp:validation}
Next, we investigate the relationship between GWD and the task performance of a \textit{NN} trained for object detection. We choose two widely used object detection datasets, the Gen1~\cite{Tournemire20arxiv} and 1 Mpx~\cite{Perot20nips} Automotive Detection Dataset, which both deal with event streams featuring labeled bounding boxes for pedestrians, cars, and other vehicles. While the former has a resolution of $304\times 240$, the latter has a resolution of $1280\times 720$. We report the GWD computed over the validation set of each dataset and then train an off-the-shelf object detection framework based on YOLOv6~\cite{yolov6_2022}, pre-trained on the Microsoft Common Objects in Context (MS-COCO) \cite{Lin14eccv}, on different input representations as in \cite{Gehrig19iccv}. To accommodate the varying number of channels, we replace the 3-channel input convolution with a $N_c=12$ channel input convolution, where $N_c$ represents the number of channels in the representation.
To show the generality of the result, we also vary the detection backbone between ResNet-50~\cite{He16cvpr}, EfficientRep~\cite{yolov6_2022} and Swin Transformer V2 \cite{liu2021swinv2}, and report results for each. \\
\textbf{Representations}: We test a range of common representations listed below. We compute 12 channels for each representation, except for the 2D Histogram, which has 2.\\
\textit{Voxel Grid}: Here, the event time window is split into $N_c$ equal, non-overlapping time windows, and then events in the same time window are aggregated by summing their polarity on a per-pixel basis using bilinear voting~\cite{Zhu19cvpr}. \\
\textit{Mixed-Density Event Stack (MDES)}: For each channel $c$, this representation selects the most recent $\frac{N_e}{2^c}$ events, where $N_e$ are the events in the time window. For each window, it aggregates the polarity at that pixel~\cite{Nam_2022_CVPR}.\\
\textit{Event Histograms}: Events in the time window are split by polarity and then summed into two channels~\cite{Maqueda18cvpr}. \\
\textit{Time Surface}: We convert events into time surfaces from~\cite{Lagorce17pami} with a decay constant of $\tau=5$ms. We then sample it at $\frac{N_c}{2}=6$ equally spaced timestamps, once for positive and negative events resulting in $N_c=12$ channels.\\
\textit{TORE}: The Time-Ordered Recent Event Volumes store event timestamps in per-pixel queues. We use queues with capacity $\frac{N_c}{2}=6$, one for positive and one for negative events, and concatenate them to $N_c=12$ channels.\\
\textit{Learned representation}: Event Spike Tensor (EST)~\cite{Gehrig19iccv} is a learnable event representation that employs a Quantization Layer featuring a trainable kernel. This approach enables the model to effectively transform raw events, optimizing their performance for a given task.\\
\textbf{Training Details} 
We adopt the training procedure from YOLOv6~\cite{yolov6_2022}. For each backbone, representation, and dataset, we train for 100 epochs, using Stochastic Gradient Descent with Nesterov momentum increasing from 0.5 to 0.83 over the first 2 epochs. We use a batch size of 32, and a Cosine learning rate schedule, starting at 0.00323 and ending at 0.000387 after 100 epochs. We adopt the classification and box regression losses in~\cite{yolov6_2022}.

\begin{figure}
    \centering
    \includegraphics[width=0.9\linewidth]{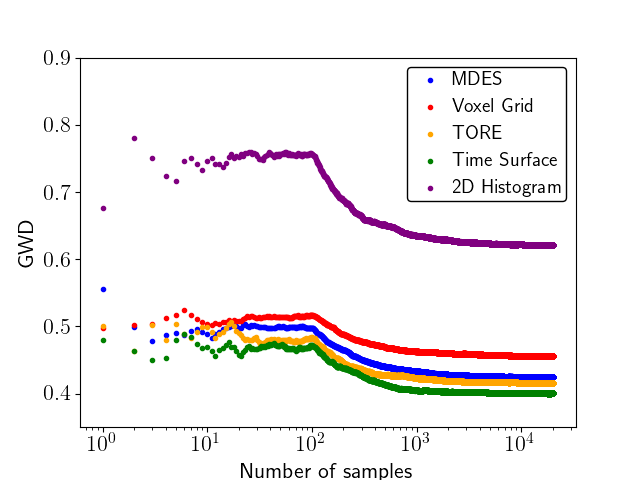}
    \vspace{-1ex}
    \caption{\gwd~for a varying number of samples from the validation set.}
    \label{fig:exp:fewer_samples}
    \vspace{-3ex}
\end{figure}

\textbf{Results}: Figs.~\ref{fig:exp:validation_task} summarizes the results of the above experiments. For both datasets, Gen1 and 1 Mpx, and all backbones, there is a clear correlation between the GWD and the task performance, i.e. task performance increases as GWD decreases. This conclusion holds for all three network backbones. In particular, MDES with a Swin-V2 detection backbone consistently achieves the highest mAP with 0.43 on Gen1 and 0.39 on 1 Mpx. It also consistently achieves the lowest GWD with 0.38 on Gen1 and 0.40 on 1 Mpx. Utilizing the learnable EST in YOLOv6 with SwinV2 backbone, we achieved a 45.31 mAP score on the Gen1 validation set and a GWD score of 0.3552, positioning EST between MDES and ERGO-12 in Fig.~\ref{fig:eyecatcher}b, confirming the expected ranking.
We also see that the Swin V2 backbone outperforms other backbones on both datasets for all tested representations.
We conclude that while the representation affects task performance, the neural network also has an influence. However, we see that the overall ranking of the representations is preserved.\\
\textbf{Using Fewer Samples}
While in the previous section, we reported the GWD over the validation set of the Gen1 and the 1 Mpx datasets, averaging this metric over such a large dataset still incurs a high computational cost and would make such a metric infeasible for optimization. 
Therefore, here we investigate if we can use smaller sample sizes to speed up the computation of GWD. In Fig.~\ref{fig:exp:fewer_samples}, we show the GWD for the representations in Sec.~\ref{sec:exp:validation} while varying the number of samples. We see that as the sample number decreases, the mean values of the metric change, but the ranking between representations is still preserved reliably after around 100 samples, and thus we use GWD$_{100}$ to optimize over representations. Below this number, the ranking of representations can fluctuate. However, we found that this happens due to a bad convergence of Eq.~\eqref{eq:GW_discrepancy}.\\
\textbf{Timing Results}: 
We time the computation of the GWD over 100 samples of the Gen1 validation set, where each sample comprises 50,000 events. We run our experiment on an AMD EPYC 7702 32-core server-grade CPU with 32 GB RAM and achieve a runtime of 15 minutes. Note that computing the GWD does not require a GPU. By contrast, training the models in Fig.~\ref{fig:exp:validation_task} requires 2 days on a single Tesla V100 GPU with 32 GB of memory, making GWD computation 192 times faster.
\input{figures_and_tables/optimization}
\subsection{Optimization of Event Representations}
\label{subsection:optimization_of_event_representations}
Here we report the results of optimizing the event representation according to the procedure in Sec.~\ref{sec:method:bayesian_optimization}. We optimize $N_c=12$ channels, and at each optimization cycle for channel $c$, the Bayesian Optimizer selects 100 configurations and then keeps the best-performing configuration.

We show the result of this optimization procedure in Fig.~\ref{fig:exp:bayesian_opt}. The left shows how the GWD decreases as new channels are added to the representation. We see that after 2 channels, our method outperforms 2D Histograms, after 6 channels we outperform Time Surfaces, after 7 channels we outperform Voxel Grids / TORE, after 9 channels we outperform Mixed-Density Event Stacks and finally, after 12 channels, we achieve a GWD$_{100}$ of 0.47. On the right, we show the different windows (columns) and measurement functions (rows) that are selected. We do not show the order since our representation is unique up to a random channel permutation. However, in the appendix, we show which features are selected at each stage. We see that all windows and all measurement functions are selected at least once, showing how our representation tries to diversify as much as possible. Moreover, timestamp-based measurements often show multiple aggregations, which we argue are necessary to replicate their complex continuous signal.  

\textbf{Comparison with State-of-the-Art}
Here we compare our method against state-of-the-art recurrent and feed-forward methods on the Gen1 and 1 Mpx test sets. We summarize the results in Tab.~\ref{tab:sota_comparison}. Recurrent methods include MatrixLSTM+YOLOv3~\cite{Cannici20eccv}, which uses a recurrent, learned input representation with a YOLOv3~\cite{Redmon16cvpr} detector, E2VID+RetinaNet~\cite{Rebecq19pami} which uses recurrent E2VID reconstructions with a RetinaNet~\cite{Lin17iccv} detector, RVT-B which uses a recurrent transformer~\cite{Gehrig22arxiv}, and RED~\cite{Perot20nips}, a recurrent CNN. ASTMNet \cite{Li22trip} utilizes asynchronous attention embedding and incorporates recurrent layers in deeper layers, claiming comparable results on both datasets. However, it is not open source. Feedforward methods are grouped into graph-based methods NVS-S~\cite{Li21iccv}, AEGNN~\cite{Schaefer22cvpr} and EAGR~\cite{eagr2022}, sparse method AsyNet~\cite{Messikommer20eccv}, spiking method Spiking DenseNet~\cite{Cordone22ijcnn}, and finally feedforward dense methods Events+RetinaNet~\cite{Perot20nips}, Events+SSD~\cite{Iacono18iros}  Events+RRC~\cite{Chen18cvprw}, and Events+YOLOv3~\cite{Jiang19icra} which use events as input, and use the RetinaNet~\cite{Lin17iccv}, SSD~\cite{Liu16eccv}, RRC~\cite{Ren17cvpr} or YOLOv3~\cite{Redmon16cvpr} detector repectively. 

\input{figures_and_tables/sota_comparison}

We compare these methods to our best-performing YOLOv6 detector from Sec.~\ref{sec:exp:validation} with a SwinV2 transformer backbone trained on various input representations. These include the ones analyzed in Sec.~\ref{sec:exp:validation}, \textit{i.e.} the 2D Histogram, Time Surface, TORE, Voxel Grid, and MDES, as well as the optimized representation ERGO-12. Note that these methods do not include data augmentation. We also trained our model with Mixup and Mosaic augmentation from~\cite{yolov6_2022}, and is indicated by an asterisk $^*$ in Tab. \ref{tab:sota_comparison}.\\
\textbf{Results} On the Gen1 dataset, we see that YOLOv6 with the SWINv2 backbone and ERGO-12 input and data augmentation outperforms all state-of-the-art methods by up to 2.9\% mAP by achieving an mAP of 0.504. The runner-up is RVT-B, which uses a recurrent vision transformer. Even without data augmentation, our network with ERGO-12 achieves 0.493, which improves the mAP by 
2.1\% compared to RVT-B. Compared to the other feed-forward methods based on YOLOv6, ERGO-12 achieves an 0.3 mAP higher mAP, the next best being YOLOv6 with Time Surface~\cite{Lagorce17pami} with 0.490.
Interestingly, as indicated in Fig. \ref{fig:eyecatcher} (b), the difference in mAP performance between ERGO-12 and Time Surface is 1.8 mAP on the validation set of the Gen1 dataset, which is substantially larger than on the test set.

On 1 Mpx, the best-performing method is ASTMNet with an mAP of 0.483, followed by RVT-B with 0.474. From the feed-forward methods YOLOv6 with ERGO-12 and data augmentation achieves the highest score with 0.406, outperforming runner-up method YOLOv6 with Time Surfaces by 2.3\% mAP. Even without augmentation, ERGO-12 achieves a 1.7\% higher score than YOLOv6 with Time Surfaces. Compared to state-of-the-art feed-forward methods, ERGO-12 with data augmentation achieves a 6.0\% higher score, the runner-up being Events+YOLOv3 with 0.346. 
On this dataset, recurrent methods are known to perform better since many sequences include stops or slow-motion scenarios. This is challenging for feed-forward methods since they are not able to maintain long-term memory. 
In general, the improvement of ERGO-12 on 1 Mpx is higher compared to Gen1.

\subsection{Object Classification}
\label{sec:exp:validation_object_classification}
As an additional task, we investigate the relationship between GWD and the object classification task performance using the \textit{ResNet-34}~\cite{He16cvpr} backbone classifier, pre-trained on ImageNet~\cite{Russakovsky15ijcv}, while changing the input convolution to be compatible with $N_c=12$ channel representations. We use the large-scale neuromorphic variant of the ImageNet dataset~\cite{Russakovsky15ijcv}, captured from an event camera that observes monitor-displayed images from ImageNet. Event sequences were recorded using the \textit{480 $\times$ 640} resolution Samsung DVS Gen3 event camera~\cite{Son17isscc}. We report the GWD computed over the validation set of the Gen1 dataset. Subsequently, we train the model on Mini N-ImageNet. We opt for GWD on the Gen1 dataset to show the generalization capability of GWD across diverse datasets. \\
\textbf{Representations}: Excluding the learned representation, we evaluate using identical representations as in object detection. Each representation is computed over 12 channels, except for the 2D Histogram, which uses two channels.\\
\textbf{Training Details}:
Adopting the methodology from the N-ImageNet study~\cite{Kim_2021_ICCV}, all inputs are resized to a \textit{224 $\times$ 224} dimension, optimizing GPU memory and inference duration. Training is initialized from scratch with a learning rate set at $3\cdot 10^-4$ and spans 100 epochs. The Adam optimizer with a Nesterov momentum of 0.9 and a weight decay of 0.0001 is used alongside a batch size of 64.\\

\noindent\textbf{Results}: Table~\ref{tab:minin_val_results} summarizes the results of the experiments. There is a clear correlation between the GWD (even on different datasets, in this case, Gen1) and the task performance, i.e. task performance increases as GWD decreases. This conclusion holds for all tested representations on the Mini N-ImageNet dataset. We obtained the following validation set accuracies: 2D Histogram (46.10\%), Time Surface (57.58\%), TORE (54.64\%), Voxel Grid (52.40\%), MDES (53.30\%), ERGO-12 (61.4\%). Their ranking is consistent with the $\text{GWD}$ ranking in Fig. 5. Therefore, the GWD can be used also for classification.\\

\input{figures_and_tables/mini_n_image_net}

\section{Conclusion}
State-of-the-art event-based deep learning methods typically need to convert raw events into dense input representations before they can be processed by standard networks. However, selecting this representation is very expensive since it requires training a separate neural network for each representation and comparing the validation scores. In this work, we circumvent this bottleneck by measuring the quality of event representations with the Gromov Wasserstein Discrepancy (GWD), which is 200 times faster to compute. We validated extensively on multiple tasks, datasets and neural network backbones that the performance of neural networks trained with a representation correlates with its GWD. We then used this metric to, for the first time, optimize over a large family of representations, revealing a new, powerful representation, ERGO-12. With it, we outperform state-of-the-art representations by 1.9 mAP on the 1 Mpx dataset and 0.3 mAP on the Gen1 dataset, two object detection benchmarks. We also exceed existing representation by 3.8\% on the tasks of classification. Moreover, we even outperform the state-of-the-art by 2.1 mAP on Gen1 and state-of-the-art feed-forward methods by 6.0 mAP on the 1 Mpx dataset. This work thus opens a new unexplored field of explicit representation optimization that will push the limits of event-based
learning methods.

\section{Acknowledgment}
This work was supported by the Swiss National Science Foundation through the National Centre of Competence in Research (NCCR) Robotics (grant number 51NF40\_185543), and the European Research Council (ERC) under grant agreement No. 864042 (AGILEFLIGHT).
\clearpage

\section{Appendix}
Here we add additional qualitative results and proofs to support the work in the main manuscript. We will refer to sections, equations, figures, and tables in the main manuscript with the prefix ``M-", while referring to those in the appendix with ``A-". 
We start by providing additional details about ERGO-12 and GWD in Sec.~A-\ref{sec:app:details_ergo12} and include two proofs regarding the robustness of \gwd~(GWD) in Sec.~A-\ref{sec:app:proofs}. Afterward, we provide more results with fewer optimized channels in Sec.~A-\ref{sec:app:fewer_optimized_channels}. Finally, we show the qualitative results of our method on the Gen1 and 1 Mpx datasets in Sec.~A-\ref{sec:app:qualitative_results}.

\subsection{Additional Details on ERGO-12 and GWD}
\label{sec:app:details_ergo12}
\textbf{ERGO-12 details:} We provide more details of our optimized representation in Fig.~A-\ref{fig:app:optimized_full}. As can be seen from the top sub-figure, we show the optimized channels in more detail than in Figure M-7. At each new step, there is a decrease in GWD, which demonstrates that additional channels reduce the distance. We calculated GWD on the Gen1 \cite{Tournemire20arxiv} validation dataset, which contained 100 samples, and plotted the results as dashed horizontal lines for chosen representations. The blue line shows the performance of the optimization process after each channel addition. We can observe that, for example, our optimized representation outperforms the Voxel Grid after seven channels and MDES after nine channels. Furthermore, we found that the optimization process initially selected the time function, which capitalizes on the high temporal resolution of event cameras to minimize GWD. Subsequently, counts and polarity were used.

In the bottom sub-figure of Fig.~A-\ref{fig:app:optimized_full}, we visualize the channels of ERGO-12 (our optimized representation after 12 channels). For visualization, we min-max normalized the channels within the range of 0-255. Each channel emphasizes different parts of the image. For instance, the last channel highlights the left edges of the pedestrian, while the seventh channel emphasizes the right part. Our optimization process enables us to capture as much information as possible at different scales and resolutions (spatial and temporal), which is highly advantageous when training with common object detectors. The optimized representation achieves an mAP of over 50\% on the Gen1 dataset, and it represents the first non-recurrent neural network architecture that scores over 40\% mAP on the 1 Mpx \cite{Perot20nips} dataset.\\
\textbf{Mathematical properties of the GWD}: The GWD introduced in \cite{peyre:hal-01322992} and used in this work does not satisfy all axioms of a distance measure and is thus not a metric. It is a generalization of the GW \textit{Distance} that is specifically designed for spaces where an L2 metric comparison is not suitable, as in this work where we compare raw events and representations. \cite{peyre:hal-01322992} showed that using KL-divergence (Eq.~9) with the kernel in Eq. 7 can effectively discard outliers, which we leverage in our work. Due to this more general formalism, the GW Discrepancy does not satisfy symmetry, or the triangle inequality (due to the KL-Divergence in Eq.~9), but ensures non-negativity, and is $0$ only for equal sets. Absolute scalability is also not satisfied (see Eq. 7), but is not a common property of distance measures.

\subsection{Invariances of the GWD for Events}
\label{sec:app:proofs}
In this section, we will go over some basic properties of the GWD for events. In particular, we will show that it is invariant to affine feature transformations, concatenation with a constant, and duplication of the features.
For clarity, we repeat here the definition of the GWD for events, following Eq. M-5: 
\begin{equation}
    L(\mathcal{E},\mathcal{F}) = \min_T \sum_{i,j,k,l}T_{ij}T_{kl}\mathcal{L}(C^e_{ik},C^f_{jl})
\end{equation}
with similarity matrices for Eqs. M-7 and M-8.
\begin{align}
C^e_{ik} = e^{\frac{-\Vert e_i - e_k\Vert^2}{2h^2\sigma_e^2}}, \quad C^f_{jl} = e^{\frac{-\Vert f_{\textbf{x}_j} - f_{\textbf{x}_l}\Vert^2}{2h^2\sigma_f^2}}
\end{align}
\begin{align}
    \sigma^2_e &= \underset{i < j}{\text{mean}}\Vert e_i - e_j \Vert^2,\quad
    \sigma^2_f &= \underset{i < j}{\text{mean}}\Vert f_{\textbf{x}_i} - f_{\textbf{x}_j} \Vert^2.
\end{align}

\textbf{Affine transformation:} We expect that if we apply an affine transformation to the event representation, the score should not change since information in the representation remains distinctive. Moreover, we do not want the GWD to be sensitive to the scale of the feature. We see that replacing representation features with 
\begin{equation}
    f^*_{\textbf{x}} = a f_\textbf{x} + b
\end{equation}
changes only the similarity matrix $C_{jl}^f$ to 
\begin{equation}
    C_{jl}^{f,*} = \exp\left(\frac{-\Vert f^*_{\textbf{x}_j} - f^*_{\textbf{x}_l}\Vert^2}{2h^2\sigma_f^{2,*}}\right).
\end{equation}
We see that the norms and data-dependent variances then transform as 
\begin{align}
\Vert f^*_{\textbf{x}_j} - f^*_{\textbf{x}_l}\Vert^2 &= a^2\Vert f_{\textbf{x}_j} - f_{\textbf{x}_l}\Vert^2\\
\sigma_f^{2,*} &= a^2\sigma_f^{2} 
\end{align}
We thus see that 
\begin{align}
    C_{jl}^{f,*} &= \exp\left(\frac{-\Vert f^*_{\textbf{x}_j} - f^*_{\textbf{x}_l}\Vert^2}{2h^2\sigma_f^{2,*}}\right)\\
    &= \exp\left(\frac{-a^2\Vert f_{\textbf{x}_j} - f_{\textbf{x}_l}\Vert^2}{2h^2a^2\sigma_f^{2}}\right)\\
    &= \exp\left(\frac{-\Vert f_{\textbf{x}_j} - f_{\textbf{x}_l}\Vert^2}{2h^2\sigma_f^{2}}\right)\\
    &=C_{jl}^f
\end{align}
which shows that the similarity matrix does not change. The minimizer of Eq. M-5 thus also does not change, which means the GWD is invariant to this affine transformation. This invariance is only possible through the use of a data-dependent variance, and thus highlights its advantage.\\
\textbf{Invariances to Concatenation}
In the case of concatenation, we consider the following transformation: 
\begin{equation}
        f^*_{\textbf{x}} = [f_\textbf{x}\Vert c_\textbf{x}]
\end{equation}
where $[.\Vert .]$ denotes concatenation, and $c_\textbf{x}\in\mathbb{R}^C$ denotes a pixel dependent additional feature. Again, we find that only the similarity matrix $C_{jl}^f$ is affected, and in particular, only the norm and variance, which become: 
\begin{align}
    \Vert f^*_{\textbf{x}_{j}} - f^*_{\textbf{x}_l}\Vert^2 = \Vert f_{\textbf{x}_{j}} - f_{\textbf{x}_l}\Vert^2 + \Vert c_{\textbf{x}_{j}} - c_{\textbf{x}_l}\Vert^2\\
    \sigma^{2,*}_f = \sigma^{2}_f + \underset{i < j}{\text{mean}}\Vert c_{\textbf{x}_i} - c_{\textbf{x}_j} \Vert^2
\end{align}
We will consider two special cases: $c_\textbf{x}=c$, a constant vector, and $c_\textbf{x}=f_\textbf{x}$ the same feature. In the first case, the additional terms above become 0, meaning that the norm does not change, and thus the metric stays the same. In the second case, the norm transforms as in the affine case, multiplying the squared norm and variance by 2. For the same reasons as before, the metric also stays the same. Generalizing this result to more general $c_\textbf{x}$ remains future work.

\subsection{Fewer optimized channels}
\label{sec:app:fewer_optimized_channels}
\begin{figure}
    \centering
    \begin{tabular}{c}
     \includegraphics[width=\linewidth]{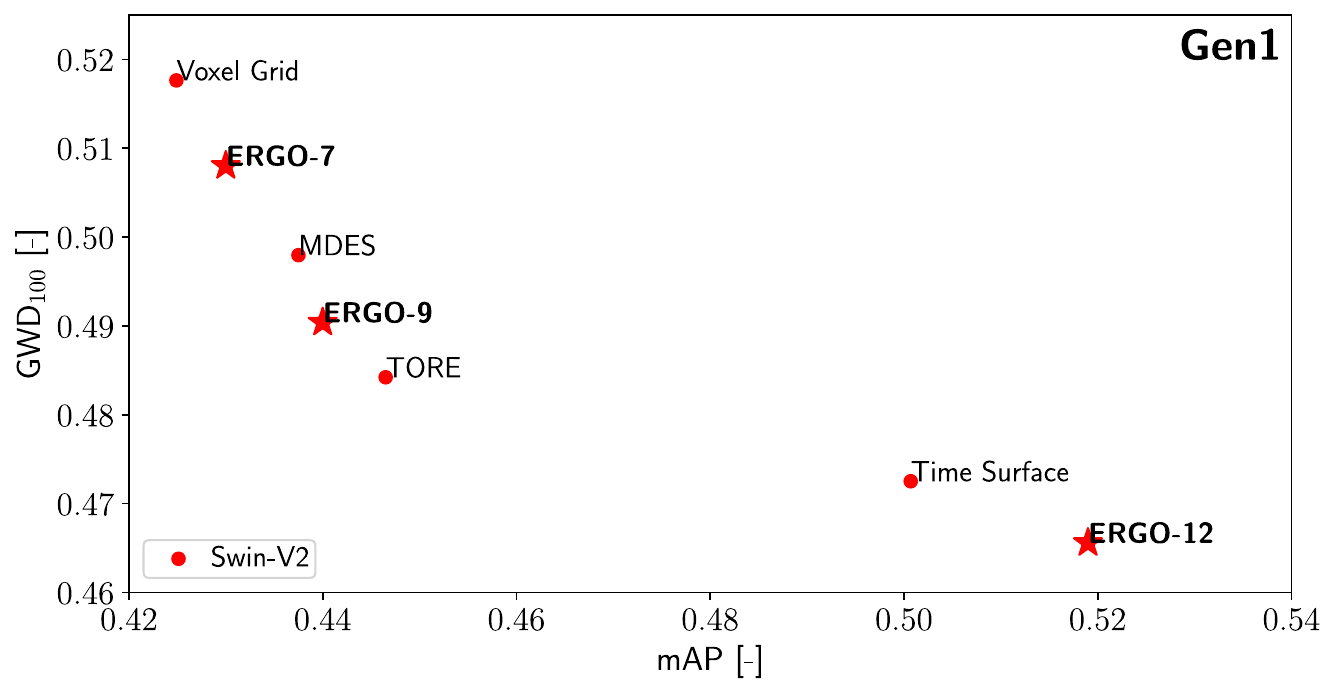}\\
         \includegraphics[width=\linewidth]{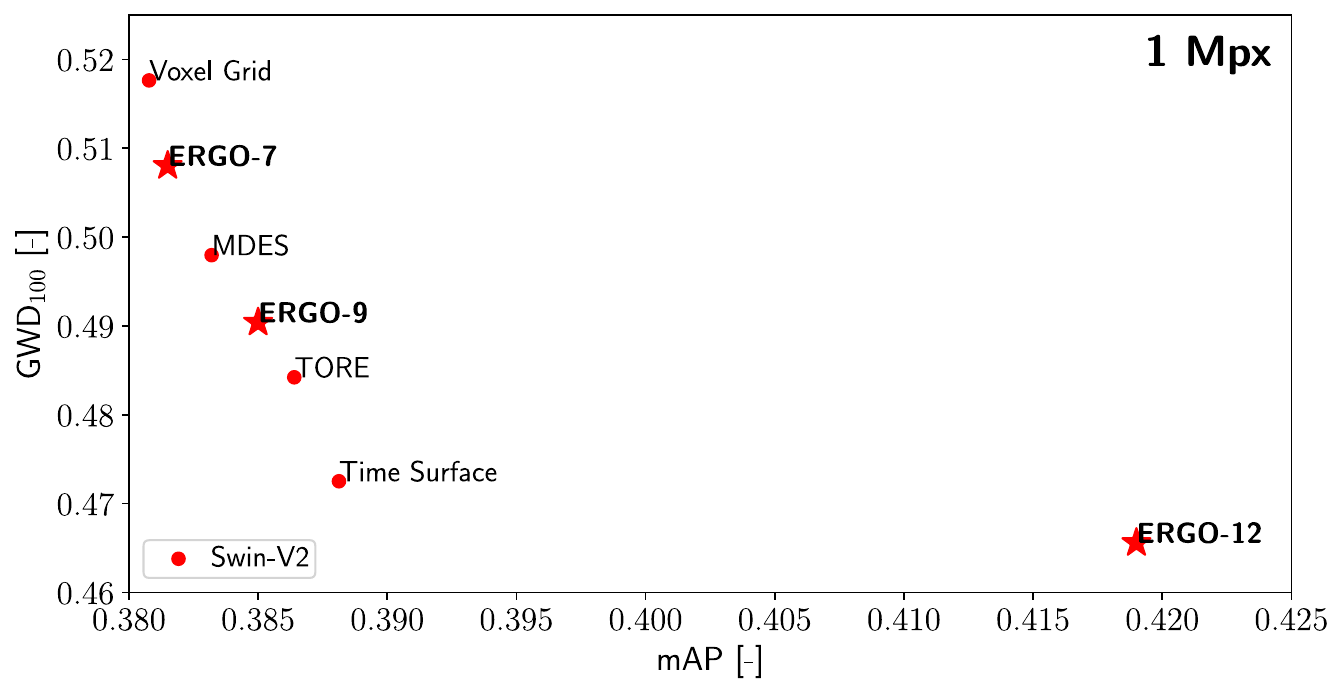}
    \end{tabular}
    \caption{Correlation of the \gwd~with the mAP (higher is better) for object detection on Gen1 \cite{Tournemire20arxiv} (top) and 1 Mpx \cite{Perot20nips} (bottom) datasets. ERGO-12, ERGO-9, and ERGO-7 represent our optimized representations with twelve, nine, and seven channels. The mAP is reported on the validation set, while the \gwd~is reported on the Gen1 validation dataset with 100 chosen samples.}
    \label{fig:varying_channels_supp}
\end{figure}

\begin{figure*}
    \centering
    \begin{tabular}{c}
     \includegraphics[width=0.9\linewidth]{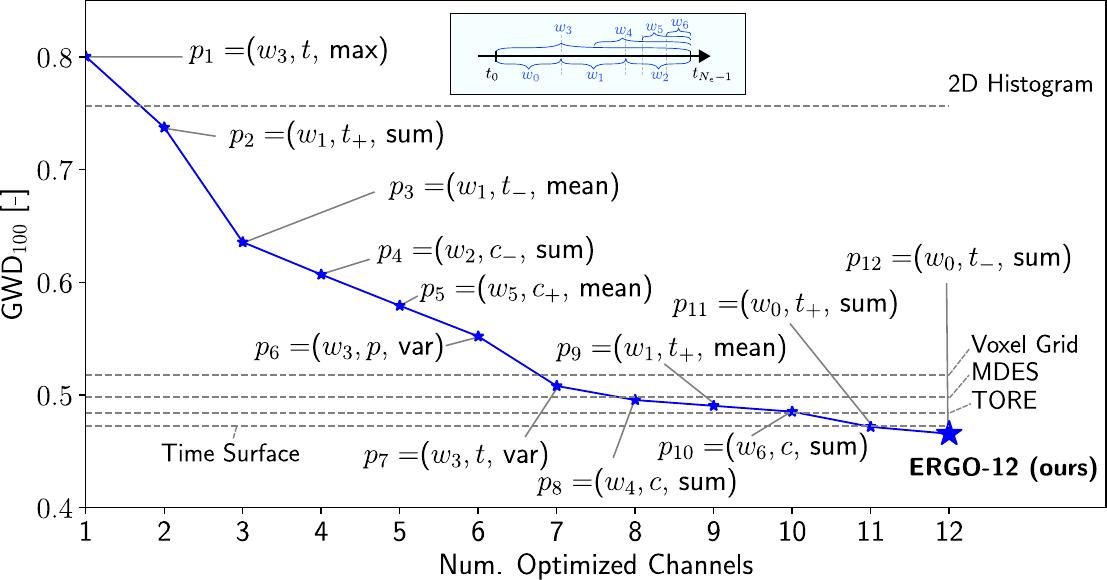}\\
         \includegraphics[width=\linewidth]{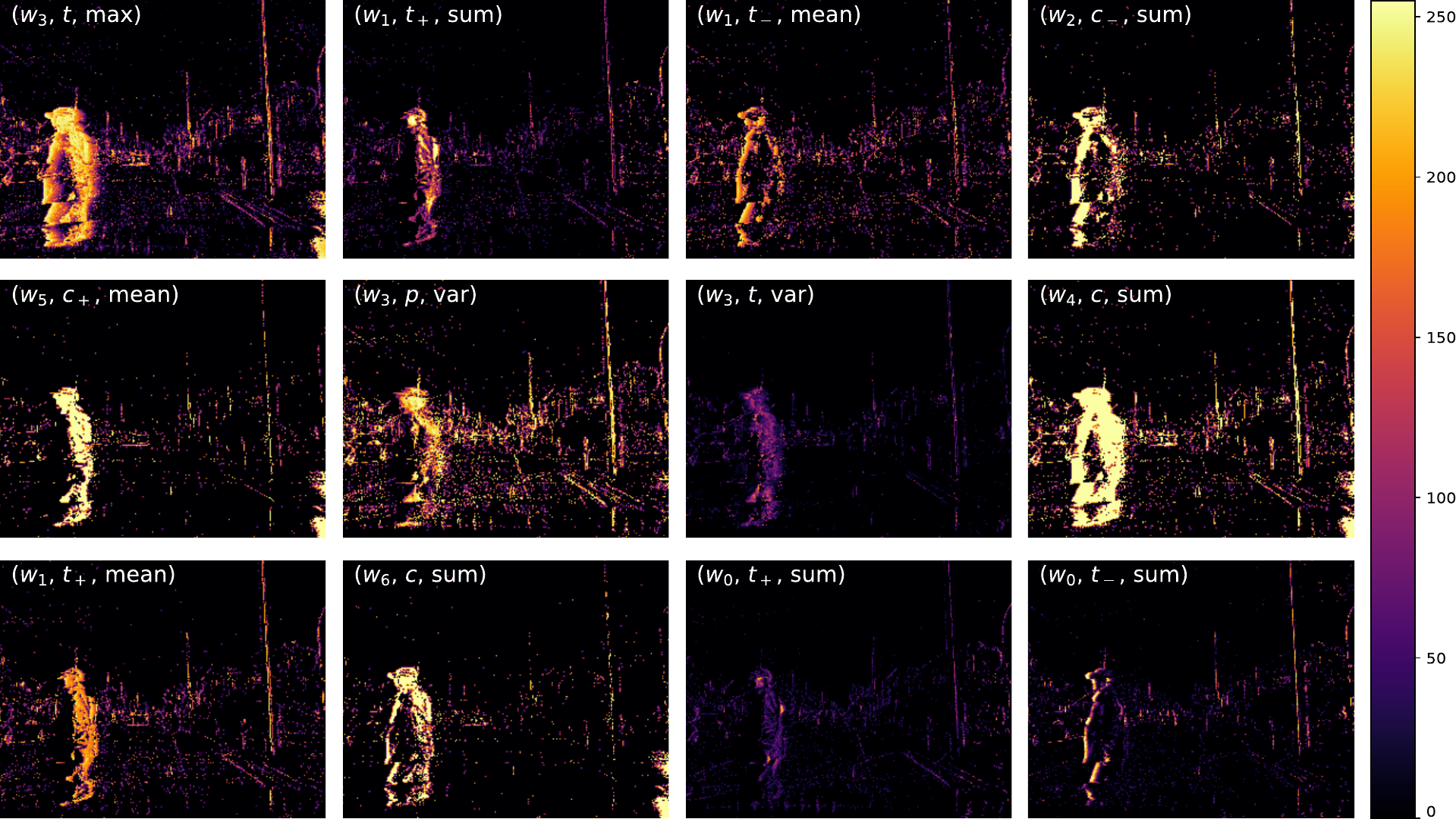}
    \end{tabular}
    \caption{Visualization of the channels of ERGO-12, min-max normalized in the range 0-255. The channels are ordered in row-major order, and the hyperparameters selected are shown in the top left of each subfigure.}
    \label{fig:app:optimized_full}
\end{figure*}

Figure \ref{fig:varying_channels_supp} depicts a correlation between the GWD (given on the x-axis, computed on the Gen1 validation dataset with 100 samples) and the task performance (mAP on object detection task). Since the Swin V2 backbone outperforms all other backbones, it is the only backbone shown in the plot, and the 2D Histogram, which is the poorest-performing method, is omitted. The results demonstrate that our optimized representation with nine and seven channels performs better than MDES and Voxel Grid, respectively, which is consistent with the findings in Figure \ref{fig:app:optimized_full}. Furthermore, we observe that the results on 1 Mpx correlate with GWD computed on the Gen1 validation dataset with 100 samples, which highlights the generalization capabilities of our approach.

\subsection{Qualitative results}
\label{sec:app:qualitative_results}
\begin{figure*}[h]
  \centering
  \setlength{\tabcolsep}{1pt}
  \resizebox{1\linewidth}{!}{%
  \begin{tabular}{cccccccc}
    \begin{turn}{90}\textbf{\quad\, Predictions}\end{turn} &
    \begin{turn}{90}\end{turn} & \includegraphics[width=\linewidth,height=0.125\textheight,keepaspectratio]{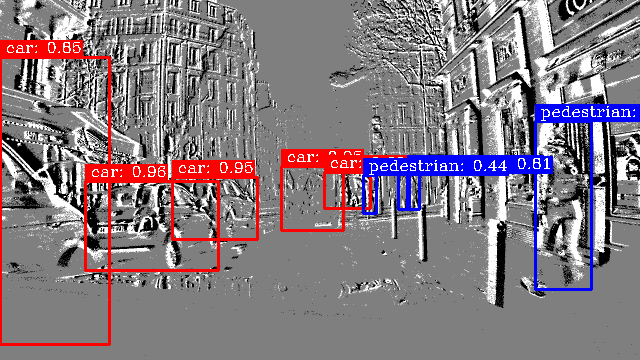} &
    \begin{turn}{90}\end{turn} & \includegraphics[width=\linewidth,height=0.125\textheight,keepaspectratio]{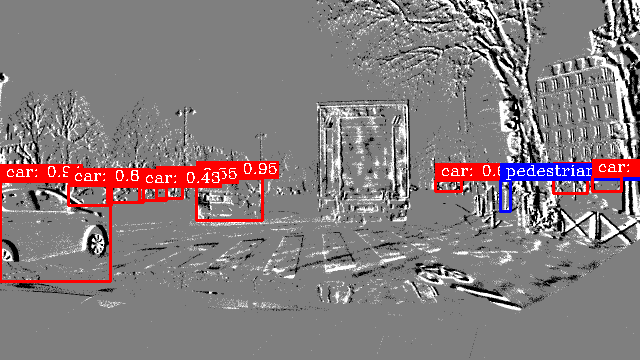} &
    \begin{turn}{90}\end{turn} & \includegraphics[width=\linewidth,height=0.125\textheight,keepaspectratio]{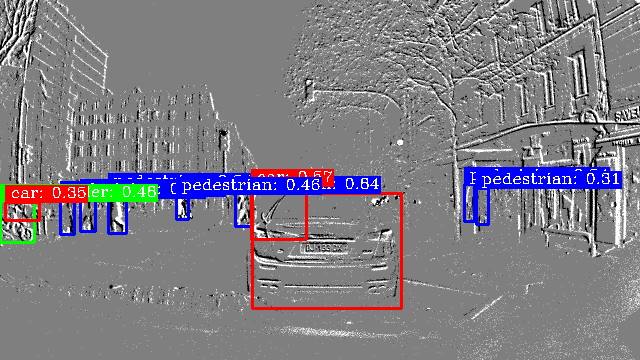} \\
    \begin{turn}{90}\textbf{\quad\,Ground-Truth}\end{turn} & 
    \begin{turn}{90}\end{turn} & \includegraphics[width=\linewidth,height=0.125\textheight,keepaspectratio]{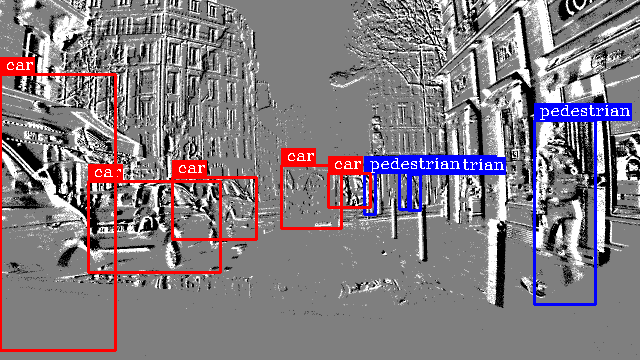} &
    \begin{turn}{90}\end{turn} & \includegraphics[width=\linewidth,height=0.125\textheight,keepaspectratio]{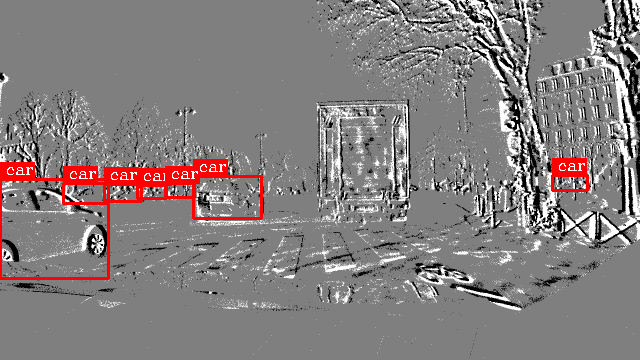} &
    \begin{turn}{90}\end{turn} & \includegraphics[width=\linewidth,height=0.125\textheight,keepaspectratio]{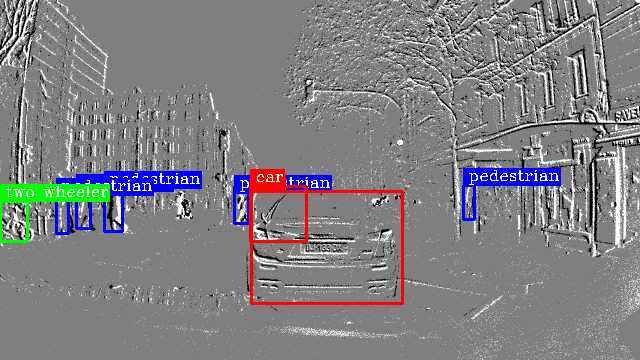} \\
  \end{tabular}}
  \caption{Qualitative results of our method with ERGO-12 input on the 1 Mpx~\cite{Perot20nips} dataset. (top row) predictions, and (bottom row) ground truth. Note that sometimes our method detects objects that do not appear in the ground truth.}
  \label{fig:gen4_pred_gt}
\end{figure*}

\begin{figure*}[h]
  \centering
\setlength{\tabcolsep}{1pt}
  \resizebox{1\linewidth}{!}{%
  \begin{tabular}{cccccccc}
    \multicolumn{2}{c}{} & \multicolumn{2}{c}{} & \multicolumn{2}{c}{} & \multicolumn{2}{c}{} \\
    \begin{turn}{90}\textbf{\quad\,\,\,Predictions}\end{turn} & \includegraphics[width=\linewidth,height=0.125\textheight,keepaspectratio]{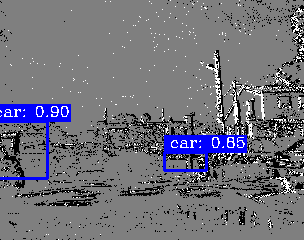} &
    \begin{turn}{90}\end{turn} & \includegraphics[width=\linewidth,height=0.125\textheight,keepaspectratio]{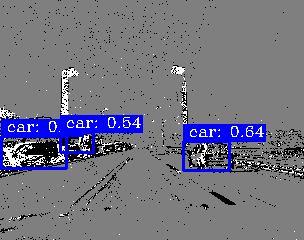} &
    \begin{turn}{90}\end{turn} & \includegraphics[width=\linewidth,height=0.125\textheight,keepaspectratio]{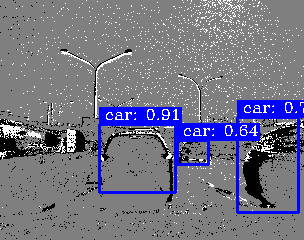} &
    \begin{turn}{90}\end{turn} & \includegraphics[width=\linewidth,height=0.125\textheight,keepaspectratio]{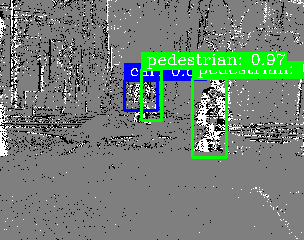} \\
    \begin{turn}{90}\textbf{\quad Ground-Truth}\end{turn} & \includegraphics[width=\linewidth,height=0.125\textheight,keepaspectratio]{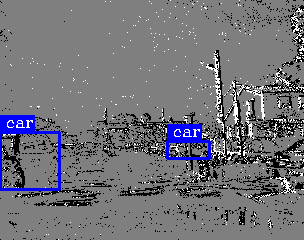} &
    \begin{turn}{90}\end{turn} & \includegraphics[width=\linewidth,height=0.125\textheight,keepaspectratio]{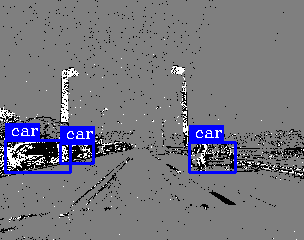} &
    \begin{turn}{90}\end{turn} & \includegraphics[width=\linewidth,height=0.125\textheight,keepaspectratio]{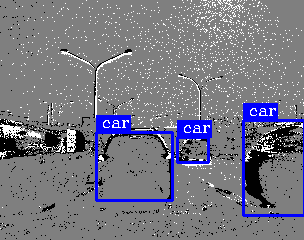} &
    \begin{turn}{90}\end{turn} & \includegraphics[width=\linewidth,height=0.125\textheight,keepaspectratio]{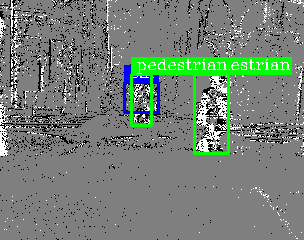} \\
  \end{tabular}}
  \caption{Qualitative results of our method with ERGO-12 input on the Gen1~\cite{Tournemire20arxiv} dataset. (top row) predictions, and (bottom row) ground truth. Note that sometimes our method detects objects that do not appear in the ground truth.}
  \label{fig:gen1_pred_gt}
\end{figure*}

We present qualitative object detection results on the 1 Mpx and Gen1 datasets in Figs. \ref{fig:gen4_pred_gt} and \ref{fig:gen1_pred_gt}, respectively. Our approach exhibits the ability to detect objects that are not present in the ground truth.

\clearpage
\clearpage

{\small
\bibliographystyle{ieee_fullname}
\bibliography{arxiv_submission}
}

\end{document}

%% file: figures_and_tables/method_overview.tex
\begin{figure*}
    \centering    
    \includegraphics[trim={0 0 0 0},clip,width=.8\linewidth]{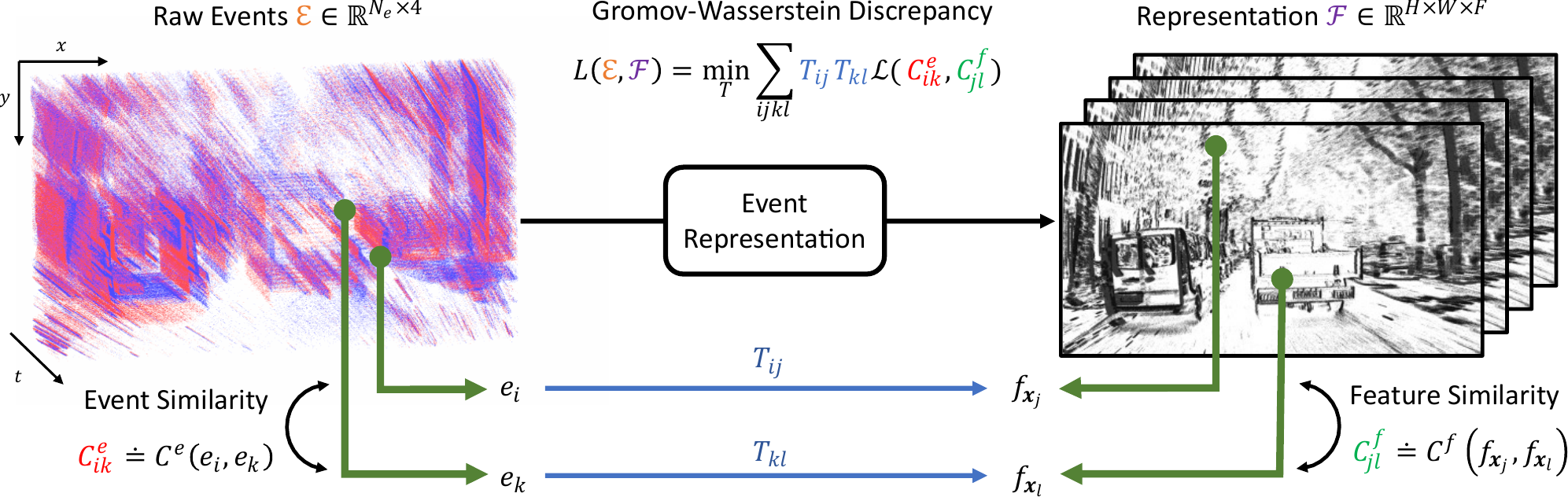}
    \caption{Overview of the \gwd~(GWD) between raw events and representations. Events $\mathcal{E}$ are converted to event representations, i.e. a set of features $\mathcal{F}$ at pixel locations $\mathbf{x}$. It is defined as the solution to an optimal transport problem which transports events pairs $(e_i, e_k)$ to feature pairs $(f_{\mathbf{x}_j}, f_{\mathbf{x}_l})$ via transport plan $T_{ij}, T_{kl}$. If the transport plan preserves the similarities $C_{ik}^e$ and $C^f_{jm}$ between event and feature pairs, this results in a low GWD.}
    \label{fig:method_overview}
    \vspace{-2ex}
\end{figure*}

%% file: figures_and_tables/hyperparameter_overview.tex
\begin{figure*}
    \centering    
    \includegraphics[width=0.8\linewidth]{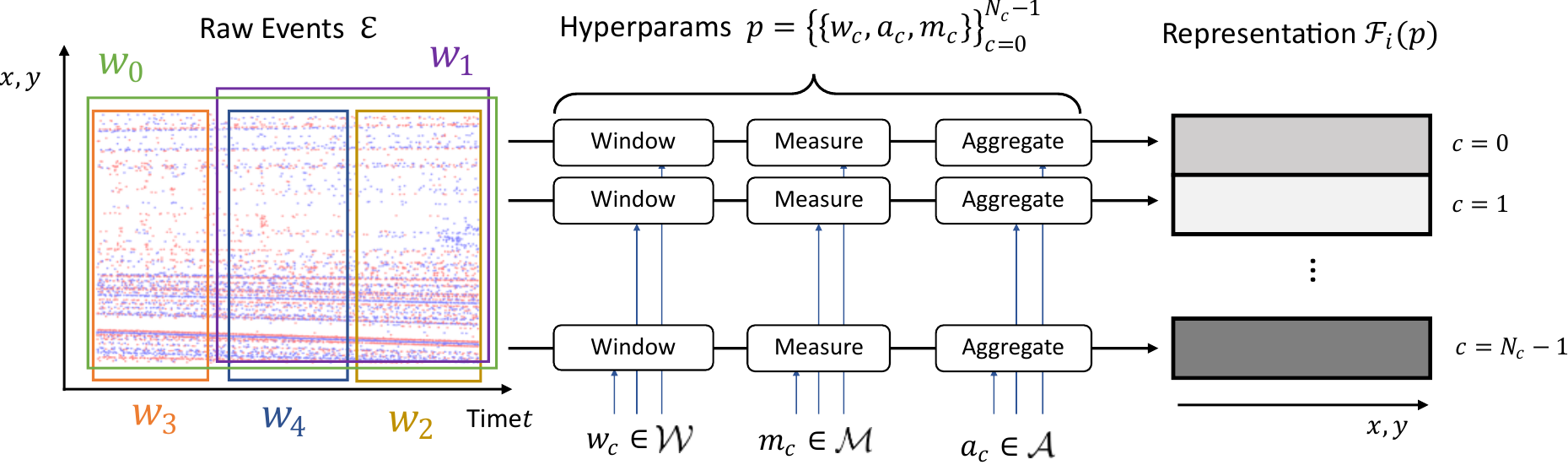}
    \vspace{-2ex}
    \caption{Overview of the hyperparameters we use to construct an event representation (right). For each channel $c$, we select one of several event time windows $w_c\in \mathcal{W}$ (in color, left), measurement functions $m_c\in\mathcal{M}$ (timestamp, polarity, positive timestamps, etc.), and aggregation functions $a_c\in \mathcal{A}$ (max, mean, sum, variance), resulting in 3$N_c$ parameters.}
    \label{fig:hyperparameter_overview}
    \vspace{-2ex}
\end{figure*}

%% file: figures_and_tables/validation_toy_example.tex
\begin{figure}
    \centering
    \begin{tabular}{c}
         \includegraphics[width=0.9\linewidth]{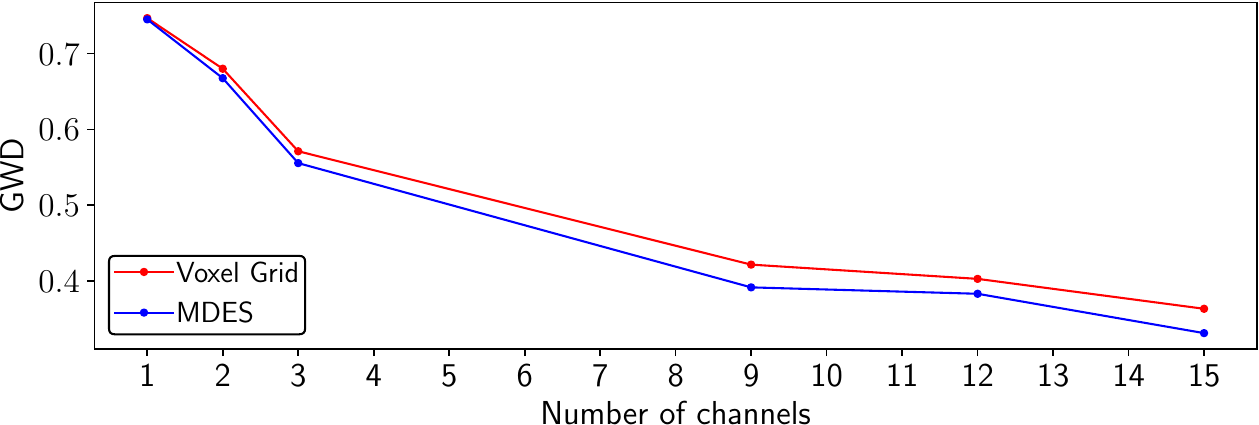}\\
         \includegraphics[width=0.9\linewidth]{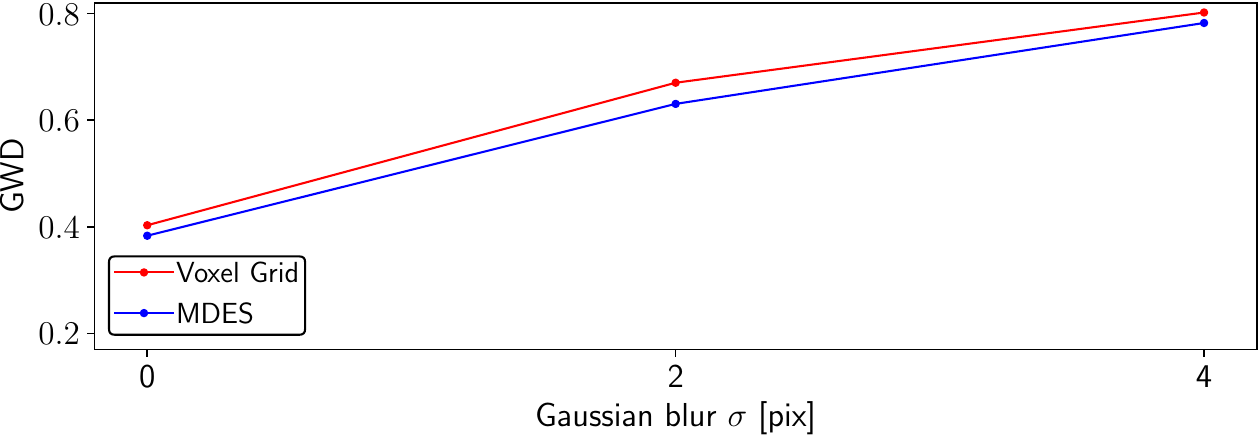}\\
    \end{tabular}
    \caption{Validation of our metric on the Gen1 validation set. (top) Our metric for Voxel Grids~\cite{Zhu19cvpr} and Mixed-Density Event Stacks \cite{Nam_2022_CVPR} with an increasing number of channels. (bottom) Effect of applying Gaussian blur with different blur kernels to the event representation. }\vspace{-2ex}
    \label{fig:exp:toy_examples}
\end{figure}

%% file: tex/validation_task.tex
\begin{figure}
    \centering
    \begin{tabular}{c}
     \includegraphics[width=0.95\linewidth]{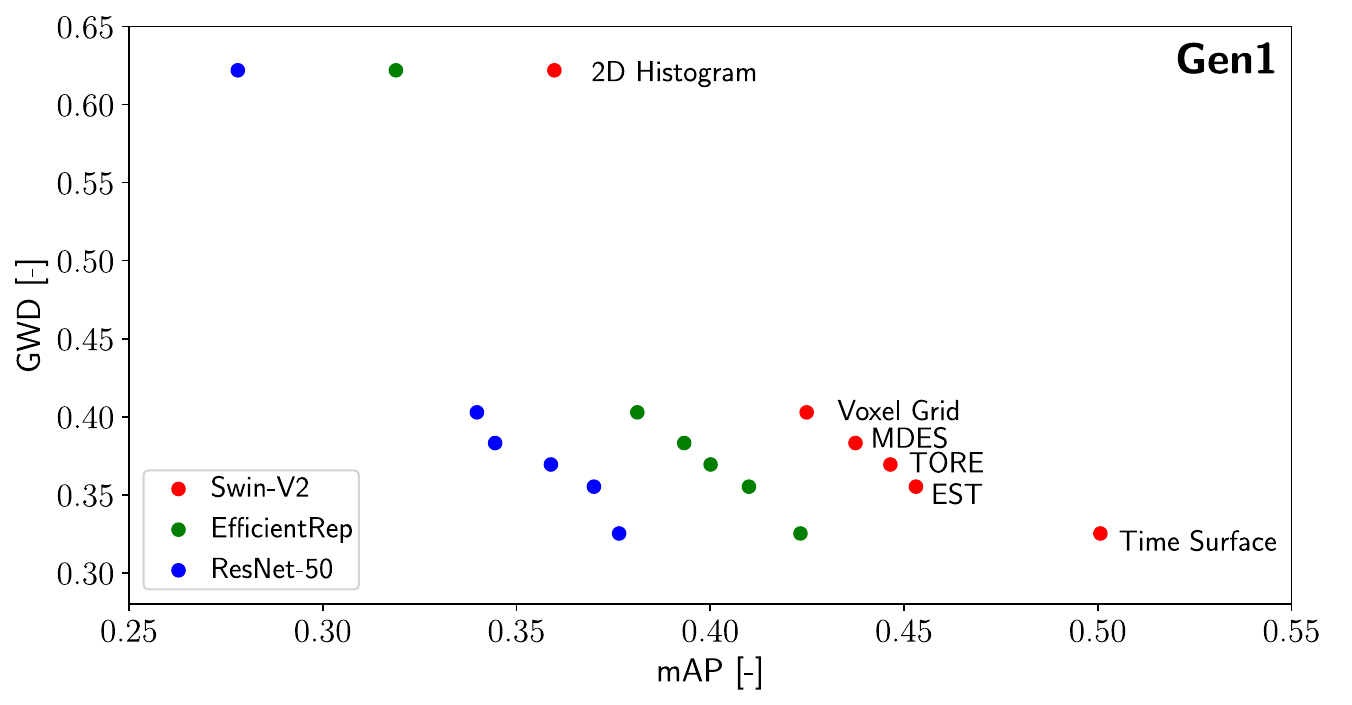}\\
         \includegraphics[width=0.93\linewidth]{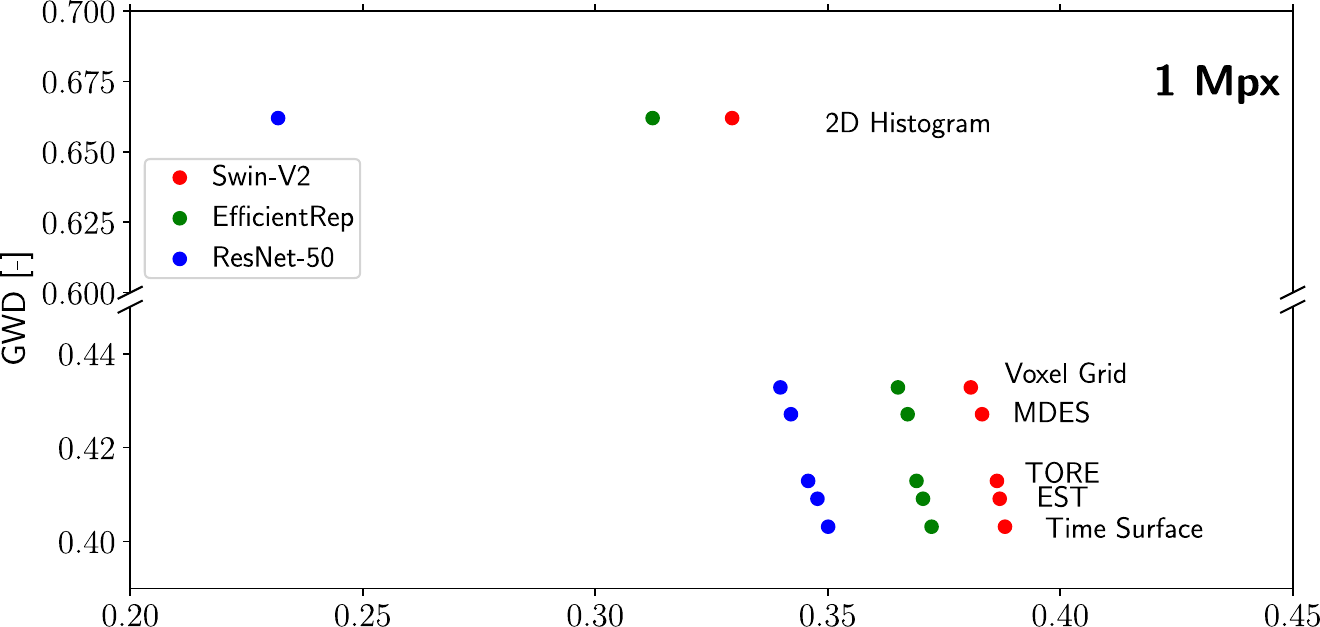}
    \end{tabular}
    \vspace{-2ex}
    \caption{Correlation of the \gwd~with the mAP (higher is better) for object detection on the Gen1~\cite{Tournemire20arxiv} (top) and 1 Mpx~\cite{Perot20nips} (bottom) datasets. Note the spliced y-axis on 1 Mpx, due to the high GWD of the 2D Histogram.}
    \label{fig:exp:validation_task}
    \vspace{-4ex}
\end{figure}

%% file: figures_and_tables/optimization.tex
\begin{figure*}
    \centering    
    \begin{tabular}{c}
         \includegraphics[width=.9\linewidth]{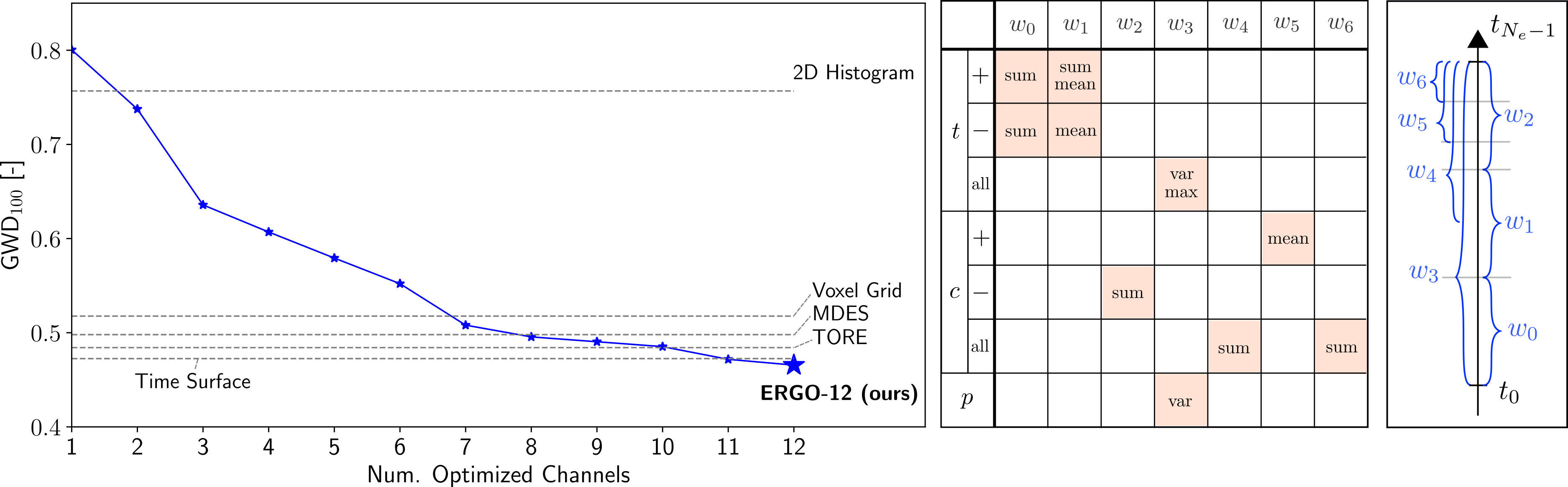}\\
    \end{tabular}
    \vspace{-1ex}
    \caption{\gwd~for 100 samples (left). At each channel, a Bayesian optimizer selects the next best hyperparameter triple. The chosen hyperparameters are broken down by window and measurement function (right). }
    \label{fig:exp:bayesian_opt}
    \vspace{-2ex}
\end{figure*}

%% file: figures_and_tables/sota_comparison.tex
\begin{table}[]
\vspace{-0.5ex}
\resizebox{\linewidth}{!}{%
\begin{tabular}{llc|c|c}
\hline
\multirow{2}{*}{\textbf{Method}} & \multirow{2}{*}{\textbf{Event Repr.}} & \multirow{2}{*}{\textbf{Recurrent}} & \multicolumn{2}{c}{\textbf{mAP}$\uparrow$} \\ \cline{4-5} 
&&& Gen1~\cite{Tournemire20arxiv} & 1 Mpx~\cite{Perot20nips} \\ \hline
MatrixLSTM+YOLOv3~\cite{Cannici20eccv}       & MatrixLSTM         & \cmark & 0.310 & -     \\
E2VID+RetinaNet~\cite{Rebecq19pami}         & Reconstructions    & \cmark & 0.270 & 0.250 \\
RED~\cite{Perot20nips}                     & Voxel Grid    & \cmark & 0.400 & 0.430 \\
RVT-B~\cite{Gehrig22arxiv}                   & 2D Histogram  & \cmark & 0.472 & 0.474 \\
ASTMNet~\cite{Li22trip}                   & Asynchronous attention embedding  & \cmark & 0.467 & \textbf{0.483} \\\hline
Events+RetinaNet~\cite{Perot20nips}        & Voxel Grid    & \xmark & 0.340 & 0.180 \\
Events+SSD~\cite{Iacono18iros}           & 2D Histogram  & \xmark & 0.301 & 0.340 \\
Events+RRC~\cite{Chen18cvprw}              & 2D Histogram  & \xmark & 0.307 & 0.343 \\
Events+YOLOv3~\cite{Jiang19icra}           & 2D Histogram  & \xmark & 0.312 & 0.346 \\
AEGNN~\cite{Schaefer22cvpr}                   & Graph         & \xmark & 0.163 & -     \\
EAGR~\cite{eagr2022}                    & Graph         & \xmark & 0.321 & -     \\
Spiking DenseNet~\cite{Cordone22ijcnn}        & Spike Train   & \xmark & 0.189 & -     \\
AsyNet~\cite{Messikommer20eccv}                  & 2D Histogram  & \xmark & 0.145 & -     \\ \hline
                        & 2D Histogram  & \xmark & 0.339 & 0.327 \\ 
                        & Time Surface  & \xmark & 0.490 & 0.383 \\
\textbf{SwinV2}       & TORE          & \xmark & 0.436 & 0.381 \\
\textbf{ + YOLOv6 (ours)} & Voxel Grid    & \xmark & 0.395 & 0.375 \\
                        & MDES          & \xmark & 0.427 & 0.378 \\
                        & ERGO-12       & \xmark & 0.493 & 0.400 \\
                        & ERGO-12      & \xmark & \textbf{0.504}* & 0.406* \\ \hline
$^*$ denotes ours with augmentation
                        \\ \hline
\end{tabular}}
\vspace{-1.5ex}
\caption{Comparison of state-of-the-art event-based object detectors on the test sets of Gen1~\cite{Tournemire20arxiv} and 1 Mpx~\cite{Perot20nips}.}\label{tab:sota_comparison}
\vspace{-4ex}
\end{table}

%% file: figures_and_tables/mini_n_image_net.tex
\begin{table}[]
\resizebox{\columnwidth}{!}{
\begin{tabular}{l|c|c|c}
\toprule
Representation & Description & \begin{tabular}[c]{@{}c@{}}\# of\\ Channels\end{tabular} & Accuracy(\%) \\ \Xhline{2\arrayrulewidth}
2D Histogram~ & \begin{tabular}[c]{@{}c@{}}Positive and negative\\ event counts\end{tabular} & 2 & 46.10 \\ \hline
\begin{tabular}[c]{@{}l@{}}Time \\ Surface~\end{tabular} & \begin{tabular}[c]{@{}c@{}}A hierarchy of\\ event-based time-surfaces\end{tabular} & 12 & 57.58 \\ \hline
\begin{tabular}[c]{@{}l@{}}TORE~ \end{tabular} & \begin{tabular}[c]{@{}c@{}}Time-ordered recent\\  event volumes\end{tabular} & 12 & 54.64 \\ \hline
Voxel Grid~ & \begin{tabular}[c]{@{}c@{}}Voxelized grid\\ with bilinear voting aggregation\end{tabular} & 12 & 52.40 \\ \hline
MDES~ & \begin{tabular}[c]{@{}c@{}}Mixed density\\ event stack\end{tabular} & 12 & 53.30 \\ \hline
ERGO-12~ & \begin{tabular}[c]{@{}c@{}}Event representation\\ from \textit{GWD} optimization\end{tabular} & 12 & \textbf{61.40} \\ \hline
\end{tabular}}
\caption{Mini N-ImageNet validation accuracy evaluated on various event representations.}
\label{tab:minin_val_results}
\end{table}

%% file: arxiv_submission.bbl
\begin{thebibliography}{10}\itemsep=-1pt

\bibitem{Alonso19cvprw}
I{\~n}igo Alonso and Ana~C Murillo.
\newblock {EV-SegNet}: Semantic segmentation for event-based cameras.
\newblock In {\em CVPRW}, 2019.

\bibitem{Baldwin2021TimeOrderedRE}
Robert~W. Baldwin, Ruixu Liu, Mohammed~Bakheet Almatrafi, Vijayan~K. Asari, and
  Keigo Hirakawa.
\newblock Time-ordered recent event (tore) volumes for event cameras.
\newblock {\em {IEEE} T-PAMI.}, 45:2519--2532, 2021.

\bibitem{bi19iccv}
Yin Bi, Aaron Chadha, Alhabib Abbas, Eirina Bourtsoulatze, and Yiannis
  Andreopoulos.
\newblock Graph-based object classification for neuromorphic vision sensing.
\newblock {\em {IEEE} Int. Conf. Comput. Vis. (ICCV)}, 2019.

\bibitem{Bi20trip}
Yin Bi, Aaron Chadha, Alhabib Abbas, Eirina Bourtsoulatze, and Yiannis
  Andreopoulos.
\newblock Graph-based spatio-temporal feature learning for neuromorphic vision
  sensing.
\newblock {\em {IEEE} Trans. Image Process.}, 29:9084--9098, 2020.

\bibitem{Cannici20eccv}
Marco Cannici, Marco Ciccone, Andrea Romanoni, and Matteo Matteucci.
\newblock A differentiable recurrent surface for asynchronous event-based data.
\newblock {\em {IEEE} Int. Conf. Comput. Vis. (ICCV)}, 2020.

\bibitem{Chen18cvprw}
Nicholas F.~Y. Chen.
\newblock Pseudo-labels for supervised learning on dynamic vision sensor data,
  applied to object detection under ego-motion.
\newblock In {\em CVPRW}, 2018.

\bibitem{chevallier_dists}
Emmanuel Chevallier, Didong Li, Yulong Lu, and David Dunson.
\newblock Exponential-wrapped distributions on symmetric spaces.
\newblock {\em SIAM Journal on Mathematics of Data Science}, 4:1347--1368, 12
  2022.

\bibitem{Cordone22ijcnn}
Loic Cordone, Benoit Miramond, and Philippe Thierion.
\newblock Object detection with spiking neural networks on automotive event
  data.
\newblock {\em Int. Joint Conf. Neural Netw. (IJCNN)}, 2022.

\bibitem{Tournemire20arxiv}
Pierre de Tournemire, Davide Nitti, Etienne Perot, Davide Migliore, and Amos
  Sironi.
\newblock A large scale event-based detection dataset for automotive.
\newblock {\em ar{X}iv e-prints}, abs/2001.08499, 2020.

\bibitem{Deng22cvpr}
Yongjian Deng, Hao Chen, Hai Liu, and Youfu Li.
\newblock A voxel graph cnn for object classification with event cameras.
\newblock {\em Conference of Computer Vision and Pattern Recognition (CVPR)},
  2022.

\bibitem{fan2021pstnet}
Hehe Fan, Xin Yu, Yuhang Ding, Yi Yang, and Mohan Kankanhalli.
\newblock Pstnet: Point spatio-temporal convolution on point cloud sequences.
\newblock In {\em (ICLR)}, 2021.

\bibitem{Gallego20pami}
Guillermo Gallego, Tobi Delbruck, Garrick Orchard, Chiara Bartolozzi, Brian
  Taba, Andrea Censi, Stefan Leutenegger, Andrew Davison, J{\"o}rg Conradt,
  Kostas Daniilidis, and Davide Scaramuzza.
\newblock Event-based vision: A survey.
\newblock {\em {IEEE} T-PAMI.}, 2020.

\bibitem{Gehrig19iccv}
Daniel Gehrig, Antonio Loquercio, Konstantinos~G. Derpanis, and Davide
  Scaramuzza.
\newblock End-to-end learning of representations for asynchronous event-based
  data.
\newblock In {\em {IEEE} Int. Conf. Comput. Vis. (ICCV)}, 2019.

\bibitem{Gehrig18eccv}
Daniel Gehrig, Henri Rebecq, Guillermo Gallego, and Davide Scaramuzza.
\newblock Asynchronous, photometric feature tracking using events and frames.
\newblock In {\em European Conference of Computer Vision (ECCV)}, pages
  766--781, 2018.

\bibitem{eagr2022}
Daniel Gehrig and Davide Scaramuzza.
\newblock Pushing the limits of asynchronous graph-based object detection with
  event cameras.
\newblock {\em arXiv}, 2022.

\bibitem{Gehrig22arxiv}
Mathias Gehrig and Davide Scaramuzza.
\newblock Recurrent vision transformers for object detection with event
  cameras.
\newblock {\em arXiv}, 2022.

\bibitem{He16cvpr}
Kaiming He, Xiangyu Zhang, Shaoqing Ren, and Jian Sun.
\newblock Deep residual learning for image recognition.
\newblock In {\em Conference of Computer Vision and Pattern Recognition
  (CVPR)}, pages 770--778, 2016.

\bibitem{Hase21apr}
Florian Häse, Matteo Aldeghi, Riley~J. Hickman, Loïc~M. Roch, and Alán
  Aspuru-Guzik.
\newblock Gryffin: An algorithm for bayesian optimization of categorical
  variables informed by expert knowledge.
\newblock {\em Applied Physics Reviews}, (8):031406, 2021.

\bibitem{Iacono18iros}
Massimiliano Iacono, Stefan Weber, Arren Glover, and Chiara Bartolozzi.
\newblock Towards event-driven object detection with off-the-shelf deep
  learning.
\newblock In {\em IROS}, 2018.

\bibitem{Jiang19icra}
Zhuangyi Jian, Pengfei Xia, Kai Huang, Walter Stechele, Guang Chen, Zhenshan
  Bing, and Alois Knoll.
\newblock Mixed frame-/event-driven fast pedestrian detection.
\newblock {\em ICRA}, 2019.

\bibitem{Kim_2021_ICCV}
Junho Kim, Jaehyeok Bae, Gangin Park, Dongsu Zhang, and Young~Min Kim.
\newblock N-imagenet: Towards robust, fine-grained object recognition with
  event cameras.
\newblock In {\em {IEEE} Int. Conf. Comput. Vis. (ICCV)}, pages 2146--2156,
  October 2021.

\bibitem{Lagorce17pami}
Xavier Lagorce, Garrick Orchard, Francesco Gallupi, Bertram~E. Shi, and Ryad
  Benosman.
\newblock {HOTS}: A hierarchy of event-based time-surfaces for pattern
  recognition.
\newblock {\em {IEEE} T-PAMI.}, 39(7):1346--1359, July 2017.

\bibitem{Lee16fns}
Jun~Haeng Lee, Tobi Delbruck, and Michael Pfeiffer.
\newblock Training deep spiking neural networks using backpropagation.
\newblock {\em Front. Neurosci.}, 10:508, 2016.

\bibitem{yolov6_2022}
Chuyi Li, Lulu Li, Hongliang Jiang, Kaiheng Weng, Yifei Geng, Liang Li, Zaidan
  Ke, Qingyuan Li, Meng Cheng, Weiqiang Nie, Yiduo Li, Bo Zhang, Yufei Liang,
  Linyuan Zhou, Xiaoming Xu, Xiangxiang Chu, Xiaoming Wei, and Xiaolin Wei.
\newblock Yolov6: A single-stage object detection framework for industrial
  applications.
\newblock {\em arXiv}, 2022.

\bibitem{Li22trip}
Jianing Li, Jia Li, Lin Zhu, Xijie Xiang, Tiejun Huang, and Yonghong Tian.
\newblock Asynchronous spatio-temporal memory network for continuous
  event-based object detection.
\newblock {\em {IEEE} Trans. Image Process.}, 31:2975--2987, 2022.

\bibitem{Li21iccv}
Yijin Li, Han Zhou, Bangbang Yang, Ye Zhang, Zhaopeng Cui, Hujun Bao, and
  Guofeng Zhang.
\newblock Graph-based asynchronous event processing for rapid object
  recognition.
\newblock In {\em {IEEE} Int. Conf. Comput. Vis. (ICCV)}, 2021.

\bibitem{Lin17iccv}
Tsung-Yi Lin, Priya Goyal, Ross Girshick, Kaiming He, and Piotr Dollár.
\newblock Focal loss for dense object detection.
\newblock In {\em {IEEE} Int. Conf. Comput. Vis. (ICCV)}, pages 2999--3007,
  2017.

\bibitem{Lin14eccv}
Tsung-Yi Lin, Michael Maire, Serge Belongie, James Hays, Pietro Perona, Deva
  Ramanan, Piotr Doll{\'{a}}r, and C.~Lawrence Zitnick.
\newblock Microsoft {COCO}: Common objects in context.
\newblock In {\em European Conference of Computer Vision (ECCV)}, pages
  740--755. 2014.

\bibitem{Liu16eccv}
Wei Liu, Dragomir Anguelov, Dumitru Erhan, Christian Szegedy, Scott Reed,
  Cheng-Yang Fu, and Alexander~C. Berg.
\newblock Ssd: Single shot multibox detector.
\newblock In {\em European Conference of Computer Vision (ECCV)}, 2016.

\bibitem{liu2021swinv2}
Ze Liu, Han Hu, Yutong Lin, Zhuliang Yao, Zhenda Xie, Yixuan Wei, Jia Ning, Yue
  Cao, Zheng Zhang, Li Dong, Furu Wei, and Baining Guo.
\newblock Swin transformer v2: Scaling up capacity and resolution.
\newblock In {\em Conference of Computer Vision and Pattern Recognition
  (CVPR)}, 2022.

\bibitem{Maqueda18cvpr}
Ana~I. Maqueda, Antonio Loquercio, Guillermo Gallego, Narciso Garc\'ia, and
  Davide Scaramuzza.
\newblock Event-based vision meets deep learning on steering prediction for
  self-driving cars.
\newblock In {\em Conference of Computer Vision and Pattern Recognition
  (CVPR)}, pages 5419--5427, 2018.

\bibitem{Messikommer20eccv}
Nico~A. Messikommer, Daniel Gehrig, Antonio Loquercio, and Davide Scaramuzza.
\newblock Event-based asynchronous sparse convolutional networks.
\newblock In {\em European Conference of Computer Vision (ECCV)}, 2020.

\bibitem{mondal2021moving}
Anindya Mondal, R Shashant, Jhony~H Giraldo, Thierry Bouwmans, and Ananda~S
  Chowdhury.
\newblock Moving object detection for event-based vision using graph spectral
  clustering.
\newblock In {\em ICCVW}, pages 876--884. IEEE, 2021.

\bibitem{Nam_2022_CVPR}
Yeongwoo Nam, Mohammad Mostafavi, Kuk-Jin Yoon, and Jonghyun Choi.
\newblock Stereo depth from events cameras: Concentrate and focus on the
  future.
\newblock In {\em Conference of Computer Vision and Pattern Recognition
  (CVPR)}, pages 6114--6123, June 2022.

\bibitem{Orchard15pami}
Garrick Orchard, Cedric Meyer, Ralph Etienne-Cummings, Christoph Posch, Nitish
  Thakor, and Ryad Benosman.
\newblock {HFirst}: A temporal approach to object recognition.
\newblock {\em {IEEE} T-PAMI.}, 37(10):2028--2040, 2015.

\bibitem{pennec:inria-00614994}
Xavier Pennec.
\newblock {Intrinsic Statistics on Riemannian Manifolds: Basic Tools for
  Geometric Measurements}.
\newblock {\em J. Math. Imaging Vis.}, 25(1):127--154, 2006.

\bibitem{PerezCarrasco13pami}
Jos\'e~A. Perez-Carrasco, Bo Zhao, Carmen Serrano, Bego{\~{n}}a Acha, Teresa
  Serrano-Gotarredona, Shouchun Chen, and Bernab\'e Linares-Barranco.
\newblock Mapping from frame-driven to frame-free event-driven vision systems
  by low-rate rate coding and coincidence processing--application to
  feedforward {ConvNets}.
\newblock {\em {IEEE} T-PAMI.}, 35(11):2706--2719, Nov. 2013.

\bibitem{Perot20nips}
Etienne Perot, Pierre de Tournemire, Davide Nitti, Jonathan Masci, and Amos
  Sironi.
\newblock Learning to detect objects with a 1 megapixel event camera.
\newblock {\em Adv. Neural Inf. Process. Syst. (NeurIPS)}, 2020.

\bibitem{peyre:hal-01322992}
Gabriel Peyr{\'e}, Marco Cuturi, and Justin Solomon.
\newblock {Gromov-Wasserstein Averaging of Kernel and Distance Matrices}.
\newblock In {\em {ICML 2016}}, June 2016.

\bibitem{Qi17nips}
Charles~R. Qi, Li Yi, Hao Su, and Leonidas~J. Guibas.
\newblock {PointNet}++: Deep hierarchical feature learning on point sets in a
  metric space.
\newblock In {\em Adv. Neural Inf. Process. Syst. (NeurIPS)}, pages 5099--5108,
  2017.

\bibitem{Rebecq19pami}
Henri Rebecq, Ren{\'{e}} Ranftl, Vladlen Koltun, and Davide Scaramuzza.
\newblock High speed and high dynamic range video with an event camera.
\newblock {\em {IEEE} T-PAMI.}, 2019.

\bibitem{Redmon16cvpr}
Joseph Redmon, Santosh Divvala, Ross Girshick, and Ali Farhadi.
\newblock You only look once: Unified, real-time object detection.
\newblock In {\em Conference of Computer Vision and Pattern Recognition
  (CVPR)}, 2016.

\bibitem{Ren17cvpr}
Jimmy Ren, Xiaohao Chen, Jianbo Liu, Wenxiu Sun, Jiahao Pang, Qiong Yan,
  Yu-Wing Tai, and Li Xu.
\newblock Accurate single stage detector using recurrent rolling convolution.
\newblock {\em Conference of Computer Vision and Pattern Recognition (CVPR)},
  2017.

\bibitem{Russakovsky15ijcv}
Olga Russakovsky, Jia Deng, Hao Su, Jonathan Krause, Sanjeev Satheesh, Sean Ma,
  Zhiheng Huang, Andrej Karpathy, Aditya Khosla, Michael Bernstein,
  Alexander~C. Berg, and Fei-Fei Li.
\newblock {ImageNet} large scale visual recognition challenge.
\newblock {\em Int. J. Com. Vis.}, 115(3):211--252, Apr. 2015.

\bibitem{SaidSalem2017RGDo}
Salem Said, Lionel Bombrun, Yannick Berthoumieu, and Jonathan~H. Manton.
\newblock Riemannian gaussian distributions on the space of symmetric positive
  definite matrices.
\newblock {\em {IEEE} Trans. Inf. Theory}, 63(4):2153--2170, 2017.

\bibitem{Schaefer22cvpr}
Simon Schaefer, Daniel Gehrig, and Davide Scaramuzza.
\newblock {AEGNN}: Asynchronous event-based graph neural networks.
\newblock 2022.

\bibitem{Sekikawa19cvpr}
Yusuke Sekikawa, Kosuke Hara, and Hideo Saito.
\newblock {EventNet}: Asynchronous recursive event processing.
\newblock In {\em Conference of Computer Vision and Pattern Recognition
  (CVPR)}, 2019.

\bibitem{Sironi18cvpr}
Amos Sironi, Manuele Brambilla, Nicolas Bourdis, Xavier Lagorce, and Ryad
  Benosman.
\newblock {HATS}: Histograms of averaged time surfaces for robust event-based
  object classification.
\newblock In {\em Conference of Computer Vision and Pattern Recognition
  (CVPR)}, pages 1731--1740, 2018.

\bibitem{Son17isscc}
Bongki Son, Yunjae Suh, Sungho Kim, Heejae Jung, Jun-Seok Kim, Changwoo Shin,
  Keunju Park, Kyoobin Lee, Jinman Park, Jooyeon Woo, Yohan Roh, Hyunku Lee,
  Yibing Wang, Ilia Ovsiannikov, and Hyunsurk Ryu.
\newblock A 640x480 dynamic vision sensor with a 9$\mu$m pixel and {300Meps}
  address-event representation.
\newblock In {\em {IEEE} Intl. Solid-State Circuits Conf. (ISSCC)}, 2017.

\bibitem{Wang19cvpr}
Lin Wang, S.~Mohammad Mostafavi~I.  , Yo-Sung Ho, and Kuk-Jin Yoon.
\newblock Event-based high dynamic range image and very high frame rate video
  generation using conditional generative adversarial networks.
\newblock {\em Conference of Computer Vision and Pattern Recognition (CVPR)},
  June 2019.

\bibitem{Zhao15tnnls}
Bo Zhao, Ruoxi Ding, Shoushun Chen, Bernabe Linares-Barranco, and Huajin Tang.
\newblock Feedforward categorization on {AER} motion events using cortex-like
  features in a spiking neural network.
\newblock {\em {IEEE} Trans. Neural Netw. Learn. Syst.}, 26(9):1963--1978,
  Sept. 2015.

\bibitem{Zhu18rss}
Alex~Zihao Zhu, Liangzhe Yuan, Kenneth Chaney, and Kostas Daniilidis.
\newblock {EV-FlowNet}: Self-supervised optical flow estimation for event-based
  cameras.
\newblock In {\em Robotics: Science and Systems (RSS)}, 2018.

\bibitem{Zhu19cvpr}
Alex~Zihao Zhu, Liangzhe Yuan, Kenneth Chaney, and Kostas Daniilidis.
\newblock Unsupervised event-based learning of optical flow, depth, and
  egomotion.
\newblock In {\em Conference of Computer Vision and Pattern Recognition
  (CVPR)}, 2019.

\end{thebibliography}
